\documentclass{article} 
\usepackage{iclr2026_conference,times}

\usepackage{amsmath,amsfonts,bm}

\def\eqref#1{equation~\ref{#1}}

\def\1{\bm{1}}

\DeclareMathAlphabet{\mathsfit}{\encodingdefault}{\sfdefault}{m}{sl}
\SetMathAlphabet{\mathsfit}{bold}{\encodingdefault}{\sfdefault}{bx}{n}

\usepackage{hyperref}
\usepackage{url}

\usepackage{amsmath}
\usepackage{empheq}
\usepackage{cases}
\usepackage{float}
\usepackage{graphicx}
\usepackage{tikz}

\usepackage{booktabs}                                   
\usepackage{multirow}                                   
\usepackage{makecell}                                   
\usepackage{tablefootnote}                              
\usepackage[symbol]{footmisc}                           
\usepackage{amsmath,amssymb}                            
\usepackage{xcolor}                                     
\usepackage{enumitem}                                   
\usepackage{subcaption}                                 
\usepackage{stfloats}                                   

\usepackage{wrapfig}                                    
\usepackage{marvosym}                                   

\usepackage{xspace}                                     
\usepackage{ulem}                                       
\usepackage{tcolorbox}                                  

\usepackage{cleveref}

\setlist[itemize]{noitemsep,topsep=0pt,parsep=1pt,partopsep=0pt,leftmargin=10pt}

\newcommand{\eat}[1]{}                                  

\newcommand{\ours}{ArtVIP\xspace}
\newcommand{\totalNum}{992\xspace}
\newcommand{\totalCat}{37\xspace}

\title{\ours: Articulated Digital Assets of \\Visual Realism, Modular Interaction, and Physical Fidelity for Robot Learning}

\author{%
    \hypersetup{pdfborder={0 0 0}}%
    Zhao Jin$^{1}$\thanks{Equal contribution. $^\dagger$Corresponding author.}
    \quad Zhengping Che$^{1*}$
    \quad Tao Li$^{1}$
    \quad Zhen Zhao$^{1}$
    \quad Kun Wu$^{1}$\\
    \textbf {Yuheng Zhang$^{1}$
    \quad Yinuo Zhao$^{1}$
    \quad Zehui Liu$^{1}$
    \quad Qiang Zhang $^{1}$
    \quad Xiaozhu Ju$^{1}$}\\
    \textbf {Jing Tian$^{2}$
    \quad Yousong Xue$^{2}$
    \quad Jian Tang$^{1\dagger}$}\\
    $^{1}$Beijing Innovation Center of Humanoid Robotics\\
    $^{2}$Beijing Institute of Architectural Design\\
    \texttt{\{mustafa.jin, z.che, jian.tang\}@x-humanoid.com}
}

\iclrfinalcopy 
\begin{document}
\maketitle

\begin{abstract}
Robot learning increasingly relies on simulation to advance complex abilities such as dexterous manipulation and precise interaction, necessitating high-quality digital assets to bridge the sim-to-real gap.
However, existing open-source articulated-object datasets for simulation are limited by insufficient visual realism and low physical fidelity, which hinder their utility for training models to master robotic tasks in the real world.
To address these challenges, we introduce \ours, a comprehensive open-source dataset comprising high-quality digital-twin articulated objects, accompanied by indoor-scene assets.
Crafted by professional 3D modelers adhering to unified standards, \ours ensures visual realism through precise geometric meshes and high-resolution textures, while physical fidelity is achieved via fine-tuned dynamic parameters.
Meanwhile, the dataset pioneers embedded modular interaction behaviors within assets and pixel-level affordance annotations.
Feature-map visualization and optical motion capture are employed to quantitatively demonstrate \ours's visual and physical fidelity, with its applicability validated across imitation learning and reinforcement learning experiments.
Provided in USD format with detailed production guidelines, \ours is fully open-source, benefiting the research community and advancing robot learning research. Our data are available at: \url{https://huggingface.co/datasets/x-humanoid-robomind/ArtVIP}.

\end{abstract}

\section{Introduction}
Embodied AI is catalyzing the transformation of robotic systems from constrained laboratory settings~\citep{billard2019trends,spong2020robot} to complex, unstructured real-world environments~\citep{brohan2023rt1roboticstransformerrealworld,zhao2023learningfinegrainedbimanualmanipulation,brohan2023rt2visionlanguageactionmodelstransfer}.
The emergence of large-scale pretrained models~\citep{zhang2023crossformer,kim2024openvlaopensourcevisionlanguageactionmodel,wang2024scalingproprioceptivevisuallearningheterogeneous} and novel learning paradigms~\citep{geminiroboticsteam2025geminiroboticsbringingai,intelligence2025pi05visionlanguageactionmodelopenworld} has ushered in a data-centric era.
In this new era, the availability of high-quality data is a critical bottleneck for developing scalable and generalizable embodied intelligence.

\begin{center}
  \includegraphics[width=0.99\linewidth]{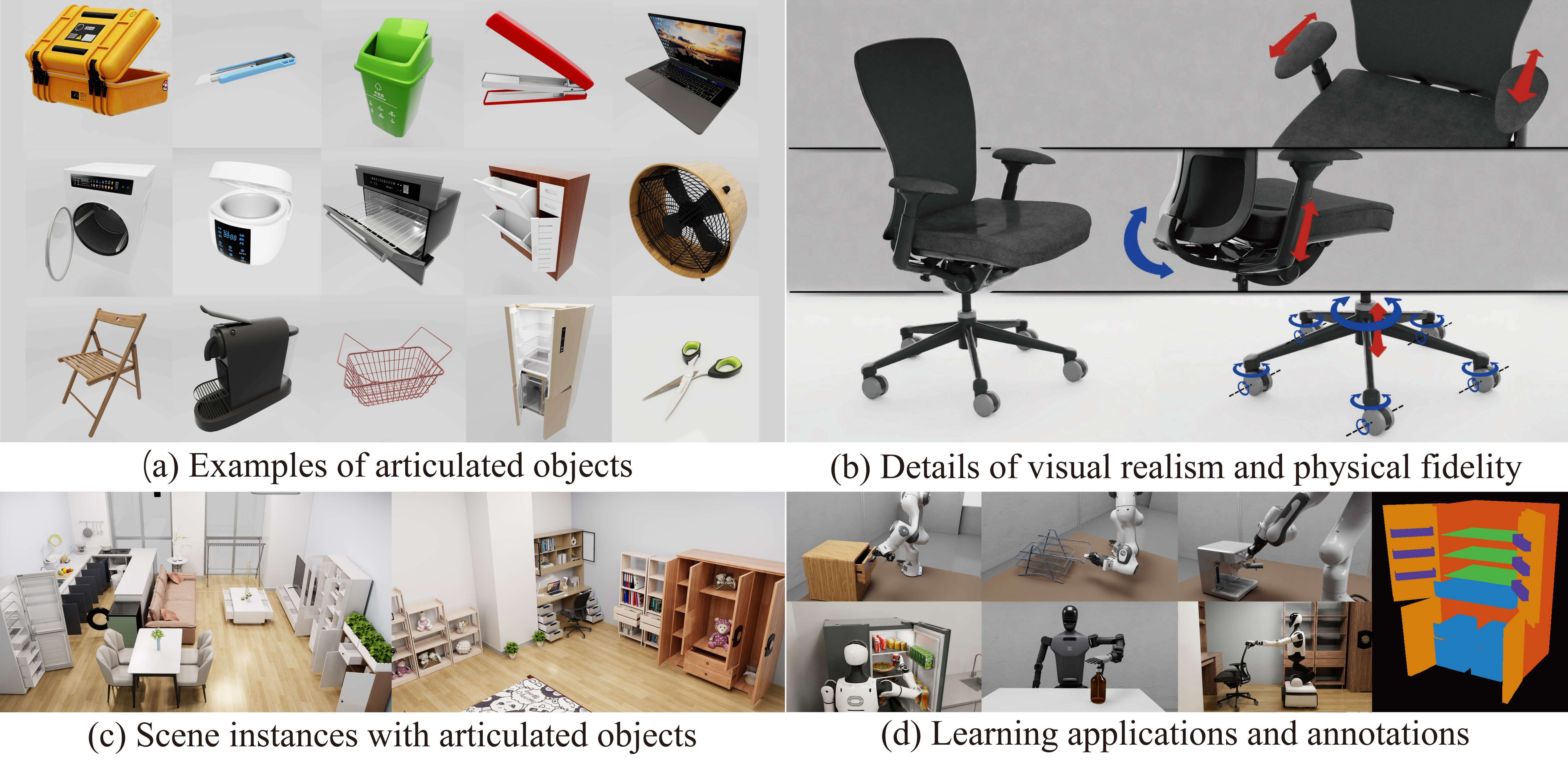}
  \captionof{figure}{\ours spans 9 categories, \totalCat\ subcategories, and \totalNum\ digital-twin articulated objects. (a) Representative assets across categories and articulation types. (b) High-fidelity physics enables realistic interactions; for example, when pushing an ergonomic chair, its casters rotate accordingly. (c) Six sim-ready scenes in which all objects support real-world–consistent interactions. (d) Pixel-level annotations and sim2real evaluations.}
  \label{fig:motivation}
\end{center}

While collecting data and deploying robots in the real world are resource-intensive and challenging to scale, simulation provides an efficient alternative for improving robot learning.
Simulation supports imitation learning by collecting unlimited and low-cost training data~\citep{wu2024robomind} and reinforcement learning by providing virtual environments~\citep{makoviychuk2021isaac,torne2024rialto}.
Meanwhile, simulations enable rapid deployment and standardized testing~\citep{ramasubramanian2022digital,do2025watch} of algorithms without concerns about hardware damage or safety issues.
Overall, simulation facilitates the exploration of innovative strategies for robot learning.

High-quality digital assets are vital to simulation for robot learning.
Simulation platforms~\citep{koenig2004design,todorov2012mujoco,ai2thor,makoviychuk2021isaac,puig2023habitat30cohabitathumans} depend on digital assets to accurately represent the real world digitally and to simulate its physical characteristics~\citep{choi2021use}.
High-quality digital assets can effectively reduce the sim-to-real gap, thereby enhancing the performance of robot learning algorithms.
For instance, digital-twin assets, which are virtual replicas created via reverse-modeling techniques, can benefit pre-deployment validation and optimization of robotic systems~\citep{straub2019replica,ramakrishnan2021hm3d}.
Moreover, high-quality digital assets can serve as training data or seed models for synthetic-asset methods such as 3D reconstruction~\citep{liu2023fewshotphysicallyawarearticulatedmesh,li2020categorylevelarticulatedobjectpose,sun2023opdmultiopenabledetectionmultiple,liu2023selfsupervisedcategorylevelarticulatedobject} and domain-randomization~\citep{dai2024acdc,ge2024behavior,torne2024rialto} techniques,
enhancing the data distribution and providing limitless diversity of objects and environments.

As robot learning turns from mastering simple tasks such as pick-and-grasp to dexterous manipulation and interaction tasks, high-quality articulated-object assets are in great demand.
Current open-source articulated-object datasets fail to meet the needs of robot learning.
For instance, PartNet-Mobility~\citep{Xiang_2020_SAPIEN} suffers from limited visual realism and insufficient physical fidelity of dynamic joints.
BEHAVIOR-1K~\citep{li2024behavior1khumancenteredembodiedai} offers better visual fidelity, but it is locked into the OmniGibson simulator~\citep{li2024behavior1khumancenteredembodiedai} and its physical parameters have not been fine-tuned. Moreover, both datasets are largely sourced from internet-searchable 3D model repositories~\citep{3dwarehouse, turbosquid} without adhering to consistent modeling standards, leading to inconsistency in quality.
Apart from using existing datasets, researchers attempt to obtain simulation assets in other ways, facing further challenges.
Retrieval-based methods~\citep{liu2024cagecontrollablearticulationgeneration,liu2024singapo} and reconstruction techniques~\citep{chen2024urdformer,Eppner2024} often inherit stylistic biases from their training data and have limited geometric generalization.
More recent pipelines~\citep{qiu2025articulateanymeshopenvocabulary3d,mandi2024real2codereconstructarticulatedobjects, le2025articulateanythingautomaticmodelingarticulated} introduce promising directions, yet face challenges such as mesh quality variance, segmentation noise, and lack of robust joint parameter tuning.

The main bottleneck for articulated object datasets lies in asset quality rather than quantity; to this end, we identify four key aspects that require careful consideration.
\begin{itemize}
  \item \textbf{Visual Realism.}
    Assets should be constructed with precise geometric meshes and high-resolution textures to ensure a photorealistic appearance.
    The amount of triangular faces should be optimized to guarantee real-time simulation performance.
  \item \textbf{Modular Interaction.}
    Assets should support interactivity (e.g., toggling a switch to turn on a light).
    These interactions should be modular to enable reuse across scenarios.
  \item \textbf{Physical Fidelity.}
    Accurate collision geometry and joint dynamics (stiffness, damping, friction) of articulated assets are essential for simulated motion to faithfully reproduce real-world kinematics and dynamics.
  \item \textbf{Simulation Friendliness.}
    Information that expands simulation usage, such as pixel-level affordance annotations and accompanying scenes, is encouraged.
    Meanwhile, open-source assets compatible with various simulation platforms and a replicable asset creation process should be provided.
\end{itemize}

To meet the mentioned requirements, we introduce \ours, a high-quality and readily deployable suite of \textit{\textbf{\underline{Art}}iculated-object digital assets with \textbf{\underline{V}}isual realism, modular \textbf{\underline{I}}nteraction}, and \textit{\textbf{\underline{P}}hysical fidelity}, 
designed to facilitate the learning and evaluation of diverse manipulation skills such as rotating, clicking, pulling, and pressing.
As illustrated in Fig.~\ref{fig:motivation}, \ours encompasses both articulated object models and complementary indoor-scene assets, all meticulously authored by professional 3D modelers under a unified asset specification to ensure consistent visual quality and realism.
Physical properties are precisely tuned to reproduce real-world dynamics, thereby enhancing the physical fidelity.
Furthermore, \ours provides pixel-level affordance annotations and uniquely embeds interaction semantics directly into the assets, enabling modular reuse and scalable behavior modeling.

In conclusion, \ours offers the following contributions:
\begin{itemize}
  \item 
    We release a collection comprising 9 categories, \totalCat subcategories, and \totalNum high-quality digital-twin articulated objects. All assets exhibit both visual realism and physical fidelity, supported by quantitative evaluations.
  \item 
    We provide digital-twin scene assets and configured scenarios integrating articulated objects within scene for immediate use.
    Extensive experiments on imitation learning, reinforcement learning, and 3D construction algorithms demonstrate the broader applicability of the assets.
  \item
    All assets are provided in the modern USD format and remain compatible with established robotics workflows via conversion to legacy formats such as URDF or MJCF.
    The detailed production process offers comprehensive guidance to facilitate community adoption and replication.
\end{itemize}

\section{Related Works}
\textbf{Simulation Platforms.}
A typical simulation platform integrates a physics engine~\citep{smith2005open,todorov2012mujoco,coumans2016pybullet,NVIDIAPhysX,tasora2016chrono} and a rendering engine~\citep{pyrender,chociej2019orrbopenairemote,rojtberg2024ogre}. Game engines~\citep{unity,unrealengine} offer similar features but do not natively support ROS~\citep{quigley2009ros,macenski2022robot} for robotics.
MuJoCo~\citep{todorov2012mujoco} and Webots~\citep{Webots} excel in simulating rigid body and multi-joint dynamics but prioritize computational efficiency over high-fidelity rendering.
Gazebo~\citep{koenig2004design}, despite its large community and robust integration with ROS, provides outdated rendering performance and exhibits lower accuracy in physical simulation.
Frameworks like AI2THOR~\citep{ai2thor}, Habitat~\citep{habitat19iccv,szot2021habitat,puig2023habitat30cohabitathumans} and ALFRED~\citep{shridhar2020alfred} are designed for mobile manipulation and instruction-following and fail to deliver precise physical interactions.
In contrast, Isaac Sim~\citep{isaacsim} offers the highest-fidelity visual rendering and leverages powerful GPU-parallel physics computation, making it well-suited for robot learning.
Other platforms, such as RoboCasa~\citep{robocasa2024} (built upon MuJoCo) and OmniGibson~\citep{li2024behavior1khumancenteredembodiedai} (built upon Isaac Sim), have become challenging to maintain. Consequently, we developed \ours specifically for Isaac Sim to capitalize on its superior rendering and physics capabilities.

\textbf{Datasets for Robot Simulation.}
Many datasets provide digital assets suitable for robot simulation. Indoor-scene assets~\citep{straub2019replica,shen2021igibson,ramakrishnan2021hm3d,li2022igibson} contribute significantly to robot navigation tasks but lack support for graphical user interface (GUI)-based editing.
Object digital asset datasets include ShapeNet~\citep{chang2015shapenet}, Objaverse~\citep{deitke2023objaverse} and other digital-twin datasets~\citep{kuang2023stanford,dong2025digitaltwincataloglargescale}.
However, these assets can only function as rigid bodies in simulations, preventing robots from performing articulated manipulation tasks with them.
Limited studies have addressed articulated object assets.
PartNet-Mobility~\citep{Xiang_2020_SAPIEN} provides 2,346 articulated-object assets across 46 categories, with many assets suffering from unsmoothed geometric surfaces, low rendering quality, and imprecise dynamic joints.
RoboCasa~\citep{robocasa2024} offers 2,508 digital assets, but only 24 are articulated objects.
BEHAVIOR-1K~\citep{li2024behavior1khumancenteredembodiedai} includes 543 articulated-object assets with improved visual fidelity, yet all assets are encrypted and accessible only through OmniGibson.
These limitations underscore the need for a high-quality, open-source articulated-object dataset.

\textbf{Articulated Objects Construction and Generation Methods.} 
Construction methods~\citep{liu2024singapo,chen2024urdformer,su2024artformer,xue2021omadobjectmodelarticulated,wang2024sm} can generate articulated objects from images and reduce labor costs. However, these methods perform reliably only on objects with simple joints, such as cabinets and desks, and produce assets with compromised visual realism. Generative methods~\citep{yang2022dsgnetlearningdisentangledstructure,liu2023meshdiffusion,long2023wonder3d,xu2023dmv3ddenoisingmultiviewdiffusion,koo2024saladpartlevellatentdiffusion} are currently limited to static rigid-body objects. These assets often exhibit distorted and implausible meshes, coupled with poor rendering quality. The absence of support for articulated objects in generative methods further limits their applicability to robot learning tasks. 

\begin{figure}[b]
  \centering
  \includegraphics[width=0.8\linewidth]{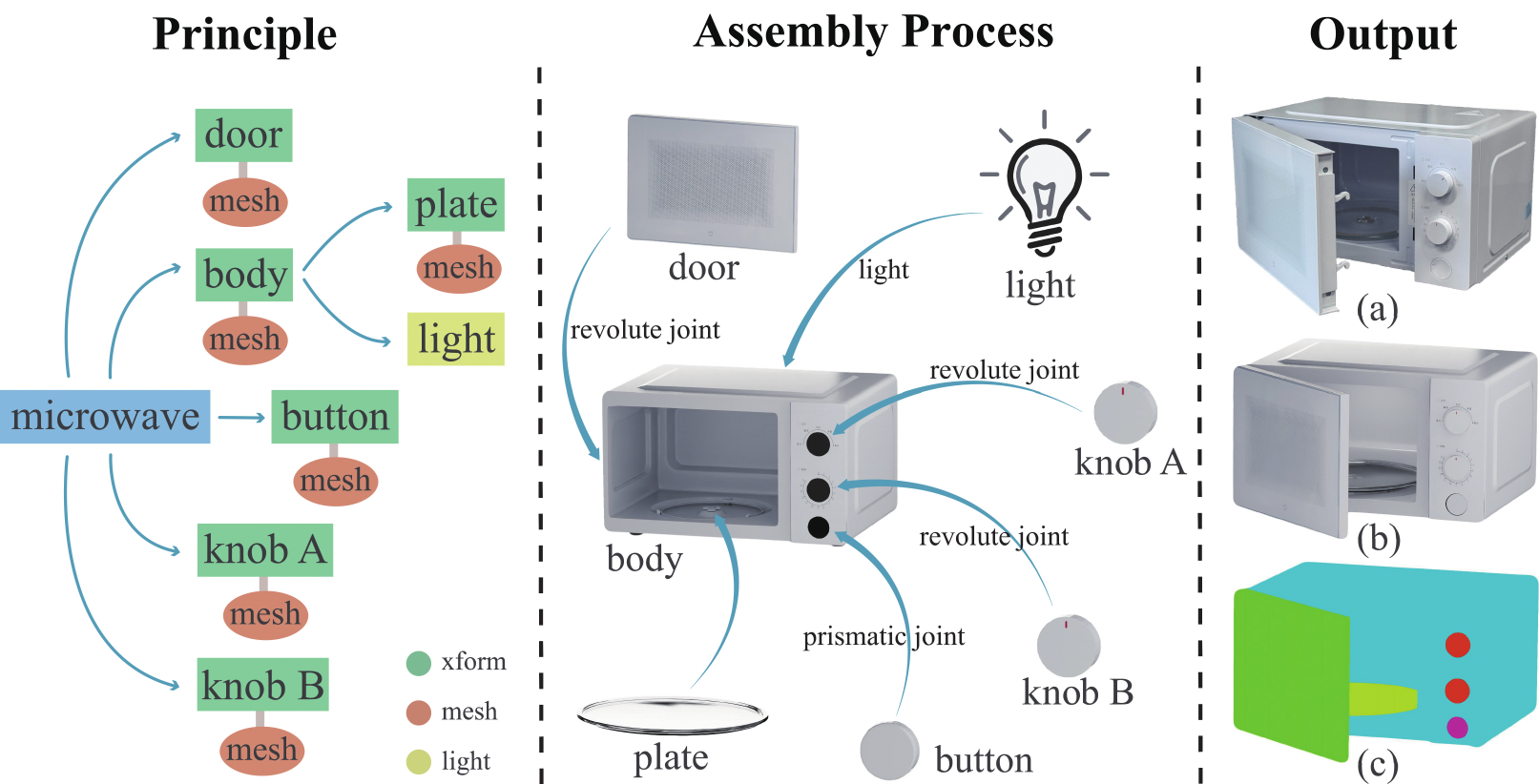}
  \caption{
    An asset in \ours. 
    \textbf{Left:} Top-down assembly principle. 
    \textbf{Middle:} Assembly process. 
    \textbf{Right:} Comparison between the real object (a) with its digital-twin (b), and annotations (c).
  }
  \label{fig:assembly}
\end{figure}

\section{\ours Collection and Methodology}
Existing datasets are largely sourced from pre-made models from public repositories. This leads to inconsistent modeling quality, disorganized part hierarchies, and non-standardized coordinate systems, all of which typically require manual preprocessing for simulation use. While current generative and reconstruction methods can easily scale up, they are still not mature enough to ensure quality. Given these constraints, we opted to prioritize fidelity over scale at this stage. 
\ours emphasizes both visual realism and physical fidelity across a comprehensive collection of articulated objects. It covers 9 categories and \totalCat subcategories, encompassing \totalNum articulated assets (see Appendix Sec.~\ref{subsec:Articulated_Objects}). Complementary sim-ready scenes (see Appendix Sec.~\ref{subsec:Scenes}) and pixel-level annotations (see Appendix Sec.~\ref{subsec:annotations}) are also provided.

\subsection{Visual Realism}
\label{visual_art}
To ensure visual realism, professional 3D modelers follow unified modeling and assembly guidelines when manually crafting articulated objects.
As shown in Fig.~\ref{fig:assembly}, we adopt a top-down mechanical modeling approach that decomposes each articulated object into three hierarchical levels: assembly, module, and mesh.
An assembly constitutes the complete functional unit, encompassing multiple modules and meshes.
Modelers first establish the assembly's base coordinate frame at the geometric center of the object's bottom surface.
Guided by the assembly's affordances, functionality, and joint locations, they partition it into rigid-body modules of the Xform type, which expose dynamic information such as transforms, velocities, and world coordinates.
Each rigid-body module contains mesh parts that provide geometric detail, visual appearance, and static physical properties, including collision shapes and mass.
Modelers follow strict rules regarding meshes, textures, and materials (see Appendix Sec.~\ref{subsec:Modeling_Standards}) to ensure visual realism.
After modeling individual meshes, they assemble them bottom-up—mesh, module, assembly—and integrate dynamic motion by connecting modules with joints (middle panel of Fig.~\ref{fig:assembly}), ensuring the asset preserves intended affordances and appearance.
Finally, for the finished asset (right panel of Fig.~\ref{fig:assembly}), each module is annotated with pixel-level labels to enable precise identification of interaction affordances.

\subsection{Physical Fidelity}
In addition to visual realism, physical fidelity plays a critical role in reducing the sim-to-real gap.
Optimized collision modeling ensures accurate rigid-body contact, improving precision in tasks such as grasping handles and other force-mediated contact scenarios.
Similarly, joint optimization yields precise joint dynamics, improving the fidelity of articulated components' motion trajectories during fine-grained operations (e.g., opening cabinet doors or pressing switches). \ours adopts the following processes.

\textbf{Collision.}
To strike a balance between physical fidelity, interaction consistency, and computational efficiency, \ours represents each mesh's collision shape using a mix of convex hulls, convex decomposition, and fine-tuned collision meshes.
For relatively regular or simple geometry, \ours relies on Isaac Sim's default convex hull generation.
When a complex mesh can be decomposed without sacrificing its affordance, 3D modelers split its collision volume into multiple primitive meshes (e.g., cubes, cylinders).
If neither a convex hull nor fine-tuned collision suffices, \ours employs Isaac Sim's built-in convex decomposition tool, which leverages mesh normals and related methods to produce accurate collision geometry.

\textbf{Joints.}
To achieve physical fidelity of dynamic joints and simulate variable joint motions in the real world, we enhance the joint drive equation~\citep{IsaacSimJointTuning} originally provided by Isaac Sim:
\begin{equation}
  \tau = K(q) \cdot \left(q - q_{\text{target}}(q)\right) + D \cdot \left(\dot{q} - \dot{q}_{\text{target}}(q)\right)
\end{equation}
where $\tau$ denotes the generalized force or torque applied to drive the joint; $q$ and $\dot{q}$ are the joint position and velocity, respectively; $D$ denotes damping; and $K$ denotes stiffness.
While this equation models basic joint motions, it does not fully capture complex joint dynamics observed in the real world.
For complex joints such as door closers and light switches, $\tau$ may vary with $q$ and $\dot{q}$.
To accommodate these cases, we parameterize $K$ and the target terms as functions of $q$, and allow dependence on $\dot{q}$ when needed.
The details are described in the Appendix Sec.~\ref{subsec:Fidelity_of_Joints}.

\subsection{Modular Interaction}
\label{modular}
\begin{figure}[!b]
  \centering
  \includegraphics[width=0.99\linewidth]{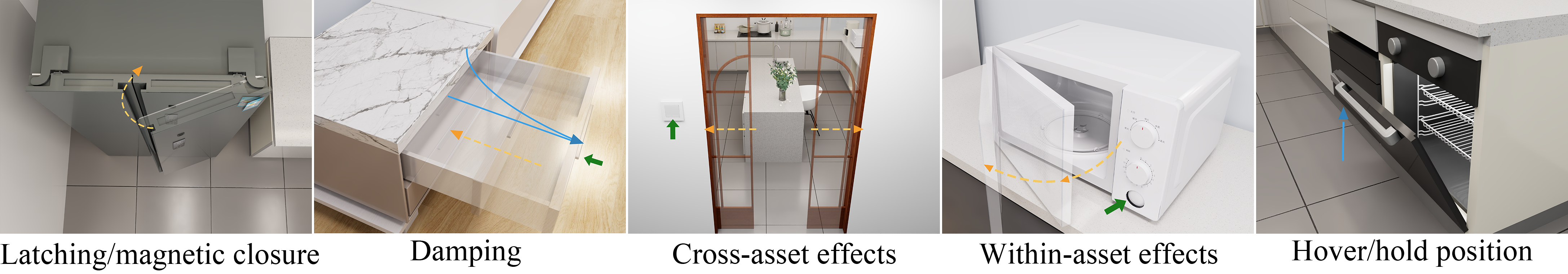}
  \caption{
  Green arrows denote applied force, yellow dashed lines indicate object motion, and blue arrows show damping. From left to right: (i) latching/magnetic closure — the door automatically closes when near shut; (ii) damping — the damping magnitude increases as the drawer is pushed in; (iii) cross-asset effects — triggering the switch opens the door; (iv) within-asset effects — pressing the microwave button opens the door and turns on the interior light; (v) hover/hold position — the oven door can hold at any angle.
  }
  \label{fig:modular}
\end{figure}
A key innovation of this work is embedding customizable behaviors directly within each asset to enable interactive functionality without writing additional code.

\textbf{Reproducing complex interactions.}
We abstract five canonical behavior primitives (Fig.~\ref{fig:modular}) for articulated objects and instantiate them across \ours, covering 394 assets and more than 900 joints.
\begin{itemize}
  \item \textit{Latching/magnetic closure:} Simulates automatic self-closing when the articulation enters a capture angle range, driven by magnetic attraction or mechanical spring/closer assemblies; once captured, a closing torque is applied until fully latched. Examples include refrigerator doors (self-closing hinge with magnetic gasket) and doors equipped with overhead closers.
  \item \textit{Damping:} Simulates sliding components and rotational hinges whose effective damping peaks near the closed position and varies smoothly along the motion, enabling gentle starts and stops. Examples include nightstand drawers, dishwashers, and cabinets.
  \item \textit{Cross-asset effects:} Simulates trigger-based coupling between distinct objects, allowing one object's state or event to drive another's behavior. Examples include button-triggered door opening and light switching.
  \item \textit{Within-asset effects:} Simulates instantaneous, mechanism-internal triggers. For example, pressing a microwave button pops the door open; similar behaviors occur in foot-pedal trash bins and height-adjustable desks.
  \item \textit{Hover/hold position:} Simulates static-friction-mediated holding in sliding or rotational joints so that, once external forces are removed, the mechanism can remain at any intermediate pose. Examples include oven doors and drawers.
\end{itemize}

\textbf{Improving asset reusability.}
Enhancing simulation development efficiency hinges on modularizing digital assets and maximizing their reusability. Our approach binds behaviors to assets at design time: researchers and artists can simply import the USD file and instantly obtain interaction affordances. This modular, reusable design reduces development overhead and accelerates algorithm iteration, allowing researchers to focus on advancing embodied AI rather than asset programming.

\section{Evaluation}
\label{eval}
We evaluate \ours along two axes: visual realism and physical fidelity, using quantitative comparisons in simulation and the real world. Comparison with generated assets are detailed in Appendix Sec.~\ref{subsec:generative_pipeline}.
\begin{figure}[!b]
  \centering
  \includegraphics[width=0.99\linewidth]{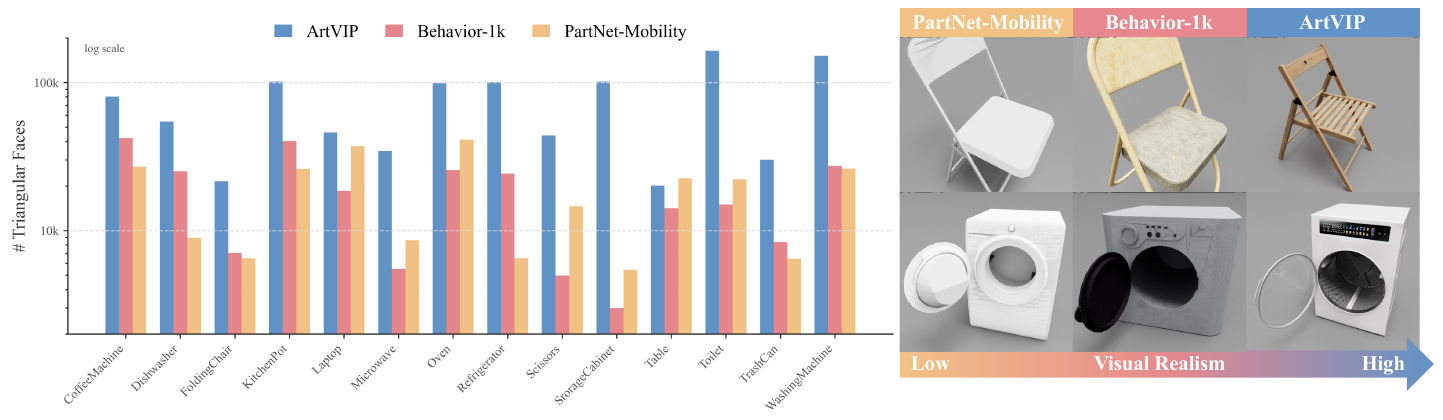}
  \caption{
    \textbf{Left:} Comparison of triangle count.
    \textbf{Right:} Rendering comparison.
  }
  \label{fig:triangles}
\end{figure}

\subsection{Visual Realism Evaluations}
We conduct a comparative analysis of \ours, BEHAVIOR-1K, and PartNet-Mobility (see Appendix Sec.~\ref{subsec:Chart_Comparison} for the detailed chart).
As shown on the right of Fig.~\ref{fig:triangles}, both BEHAVIOR-1K and PartNet-Mobility exhibit distorted geometry and implausible appearance. In addition, we quantify geometric detail via triangle count, evaluate reconstruction performance, and visualize feature distributions to assess visual realism. 

\textbf{Geometric Detail.}
Meshes built from densely triangular faces preserve the core geometric detail.
A high count of triangular faces improves surface smoothness and minimizes faceting.
The left of Fig.~\ref{fig:triangles} illustrates the comparison results on object categories that appear in all three datasets, demonstrating the rich geometric detail in \ours.
More analysis and relative profiling are in the Appendix Sec.~\ref{subsec:Visual_Comparison}.

\begin{figure}[t]
  \centering
  \begin{tikzpicture}
      \node[anchor=south west,inner sep=0] (image) at (0,0) {\includegraphics[width=\textwidth]{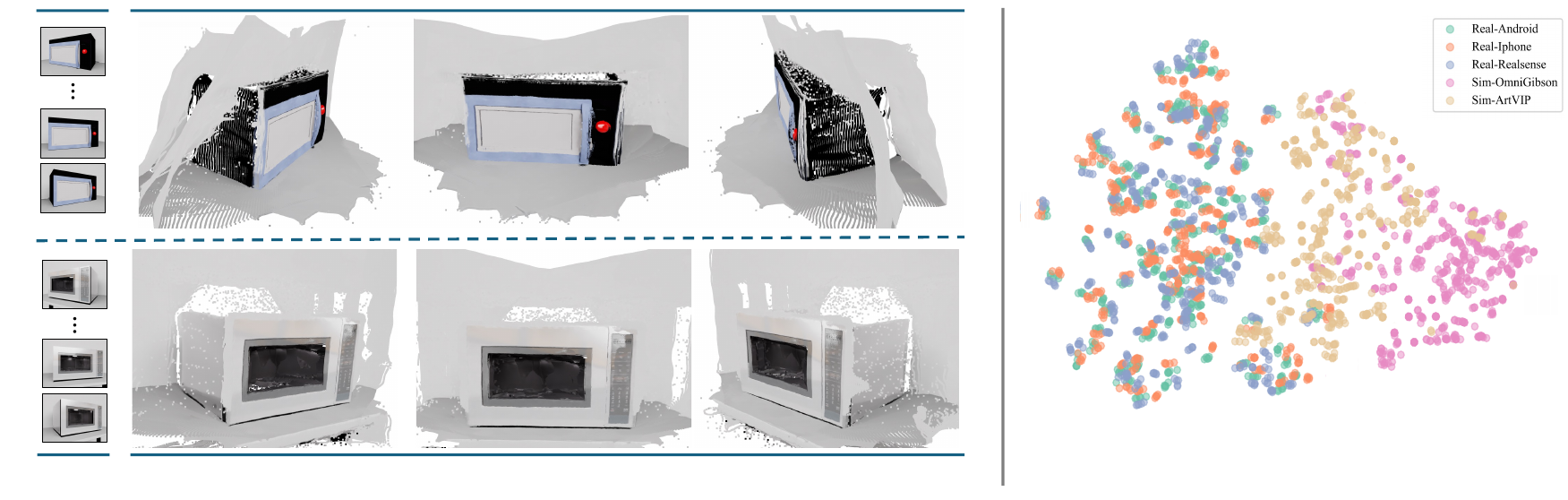}};
      \begin{scope}[x={(image.south east)},y={(image.north west)}]
          \node[rotate=90, font=\scriptsize] at (0.01,0.75) {OmniGibson}; 
          \node[rotate=90, font=\scriptsize] at (0.01,0.28) {ArtVIP};     
          \node[font=\normalsize] at (0.053,0.035) {\fontsize{8}{9.6}\selectfont Sampling};
          \node[font=\footnotesize] at (0.35,0.035) {\fontsize{8}{9.6}\selectfont Reconstruction};
          \node[font=\footnotesize] at (0.83,0.035) {\fontsize{8}{9.6}\selectfont Feature Distribution}; 
      \end{scope}
  \end{tikzpicture}
  \caption{
    \textbf{Left:} Reconstruction of a microwave. 
    OmniGibson yields poor results due to weak visual appearance, while \ours enables better reconstruction via more realistic details. 
    \textbf{Right:} CLIP-based~\citep{radford2021learning} feature distribution.
    Each color denotes a data source and \ours features align more closely with real-world data.
  }
  \label{figx}
\end{figure} 

\textbf{Reconstruction Performance Evaluation.}
To assess differences in reconstruction quality across data assets, we conducted experiments using VGGT~\citep{wang2025vggt}, a widely adopted method that has demonstrated strong generalization in real-world reconstruction tasks.
Using identical multi-view sampling strategies on the OmniGibson and ArtVIP assets, we generated reconstruction inputs, with results shown on the left portion of Fig.~\ref{figx}.
Reconstructions from \ours assets exhibit higher structural fidelity and finer detail preservation compared to those from OmniGibson. 
This suggests that \ours's more realistic geometry and material representation enhance the quality and compatibility of sampled images for reconstruction tasks.
The results underscore the role of high-fidelity assets in supporting viewpoint diversity and accurate structure recovery.

\textbf{Feature Distribution Visualization Analysis.}
To verify the visual realism of \ours assets, we randomly sampled 100 3D models and selected corresponding or semantically similar objects from OmniGibson and the real world for comparison. 
Real-world images were captured using three devices (an Android phone, an iPhone, and an Intel RealSense D435) under multi-view settings. 
In Isaac Sim, we rendered samples of the ArtVIP and OmniGibson assets using matched camera viewpoints to ensure consistency across domains.
We applied t-SNE~\citep{van2008visualizing} to visualize the extracted CLIP~\citep{radford2021learning} features.
As shown on the right portion of Fig.~\ref{figx}, \ours features align more closely with real-world data, indicating higher consistency in visual semantics, texture, and material.
This fidelity enhances the value of \ours for simulation-to-reality transfer in downstream tasks.

\begin{figure}
  \centering
  \includegraphics[width=0.97\linewidth]{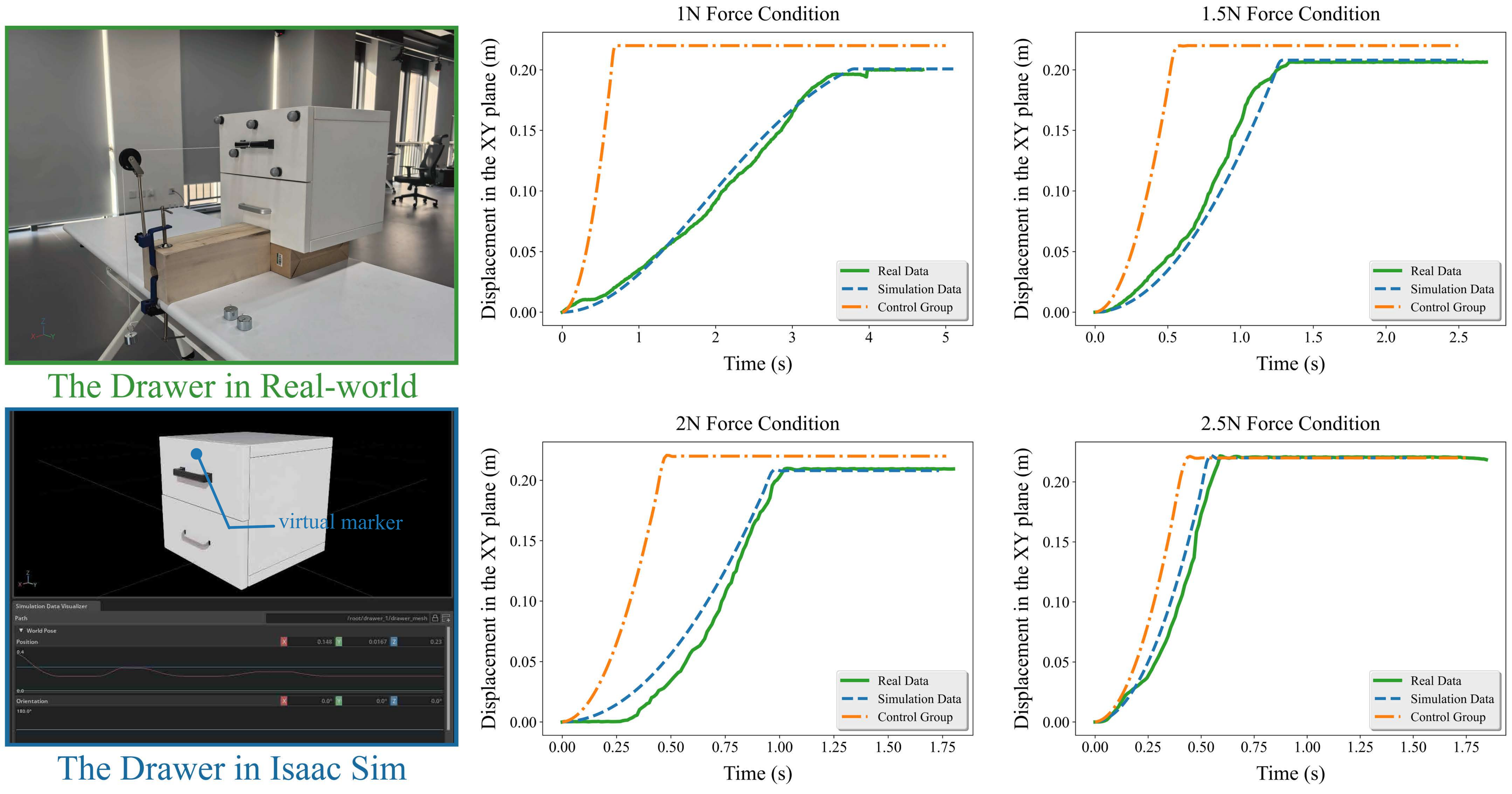}
  \caption{
    \textbf{Left:} Digital-twin asset examples in real-world and simulation.
    \textbf{Right:} Analysis of the drawer's displacement driven by different forces.
  }
  \label{fig:mocap}
\end{figure} 

\subsection{Physical Fidelity and Interaction Evaluations}
To demonstrate the physical fidelity of joint motion within articulated objects, we employed an optical tracking system ($0.1$ mm spatial resolution and $90$ Hz sampling rate) to record motion trajectories of joints on real-world objects.
These recordings were compared with the joint motions of their corresponding digital-twin articulated objects in simulation to evaluate the discrepancy between simulated and real-world joint behavior. 
We test in a common scenario where joint motion triggered by external force.
More setting descriptions and evaluation results are described in the Appendix Sec.~\ref{subsec:Physical_Evaluations}.

As shown in Fig.~\ref{fig:mocap}, in the real-world experiment, horizontal pulling forces of $1$ N, $1.5$ N, $2$ N, and $2.5$ N were applied to the drawer by suspending calibrated weights from the end of the fixed pulley system, ensuring consistent force direction.
The drawer's displacement in the XY plane was recorded in real time.
In the simulation environment, two configurations were evaluated:
one with default joint parameters and the other with optimized parameters.
Both were subjected to the same force configuration as the real-world setup, and the spatial trajectories of the drawer's keypoints were tracked.
The close agreement between the displacement obtained from simulation and real-world experiments, as shown in the right of Fig.~\ref{fig:mocap}, demonstrates the physical fidelity of the joints in \ours.

\section{Applications}
\label{application}
To further verify the capability of \ours in supporting downstream robotic learning tasks, we conducted extensive experiments in both the real-world and simulated environments following two primary paradigms in robotic learning: Imitation Learning and Reinforcement Learning.

\subsection{Imitation Learning in Real World Environments}

\begin{figure}[t]
  \centerline{\includegraphics[width=0.99\linewidth]{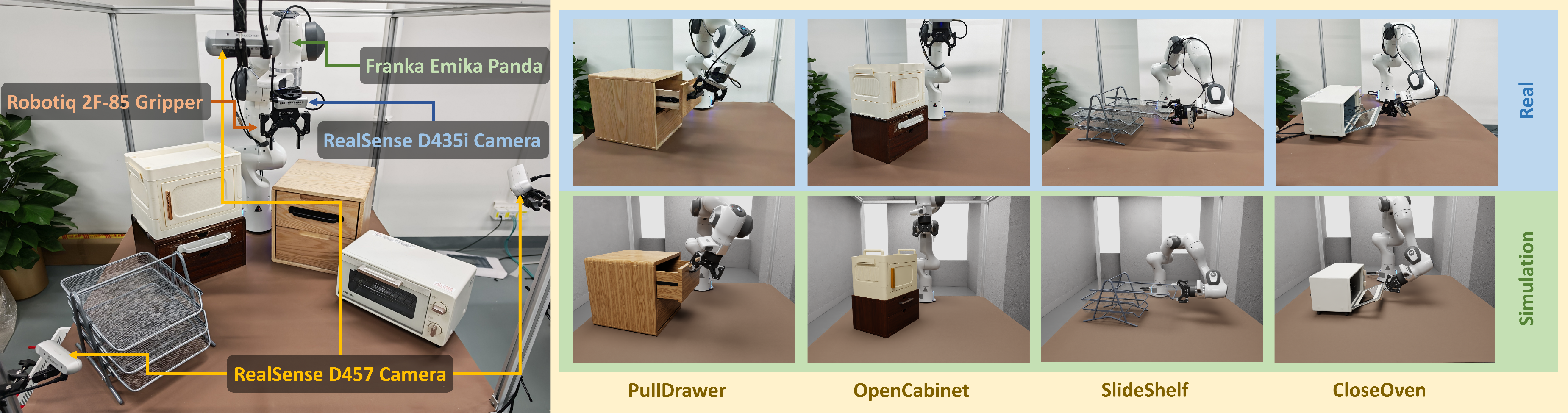}}
  \caption{Experimental Setup. We conducted 4 real-world tasks for imitation learning.}
  \label{fig:il_hardware_setup}
\end{figure}

\textbf{Experimental Setup.}
As illustrated in Fig.~\ref{fig:il_hardware_setup}, we used a Franka robotic arm equipped with a Robotiq 2F-85 gripper and four RealSense cameras to create the real-world experimental environment. 
These cameras include three external RealSense D457 cameras (placed on the left, right, and top of the table) and one hand-eye RealSense D435i camera mounted at the wrist of the robotic arm. 
For simulation, we used Isaac Sim and replicated this real-world setup, including the Franka robotic arm, the operating table, camera settings, and the manipulated objects from \ours.
We constructed the simulated scene to match the real-world experiment environment as closely as possible.

\textbf{Task Design and Data Collection.}
As shown in Fig.~\ref{fig:il_hardware_setup}, we design four challenging articulated-object manipulation tasks:
(1) \textbf{PullDrawer},
(2) \textbf{OpenCabinet},
(3) \textbf{SlideShelf}, and
(4) \textbf{CloseOven}.
These tasks demand precise and flexible motions, including rotation, angled pushing, and horizontal translation (see Appendix Sec.~\ref{subsec:IL_APP}).
Data was collected via teleoperation in both real and simulated environments, where articulated objects were randomly placed within a predefined workspace and human operators completed each task.
For each task, we gathered 100 successful trajectories in the real world and 100 in simulation.
Each trajectory includes RGB streams from four camera viewpoints and full proprioceptive robot states (e.g., joint positions) throughout execution.

\textbf{Imitation Learning Algorithm.}
We used two canonical imitation learning baselines, Action Chunking Transformer (ACT)~\citep{zhao2023learningfinegrainedbimanualmanipulation} and Diffusion Policy (DP)~\citep{chi2023diffusion}, to train the robotic policies for the articulated object manipulation task (more details in Appendix Sec.~\ref{subsec:IL_APP}).

\textbf{Experimental Results on Imitation Learning.}
For each of the four articulated-object manipulation tasks, we trained ACT and DP under the following dataset settings: \textbf{(1) Real-Only (RO)}: 100 real-world trajectories; \textbf{(2) Sim-Only (SO)}: 100 simulated trajectories; \textbf{(3) Real–Sim–Mixed (RSM100+10/20/50/100)}: 100 real-world + {10, 20, 50, 100} simulated trajectories. For each experiment, we trained ACT and DP for 50k gradient descent iterations with three different random seeds, and evaluated the final checkpoint from each run with 60 rollouts to compute per-task success rates.

\begin{table}[t]
\caption{Success rates of ACT and DP across dataset settings: RO (real-only), SO (sim-only), and RSM variants for all tasks.}
\label{tab:il_results}
\begin{center}
  \resizebox{0.8\columnwidth}{!}{
    \begin{tabular}{cccccc}
    \toprule
       Method & Dataset & PullDrawer & OpenCabinet & SlideShelf & CloseOven \\
    \midrule
       \multirow{6}{*}{ACT~\citep{zhao2023learningfinegrainedbimanualmanipulation}} 
          & RO           & 64\% & 34\% & 27\% & 58\% \\
          & SO           & 39\% & 12\% & 13\% & 23\% \\
          & RSM100+10    & 64\% & 36\% & 26\% & 59\% \\
          & RSM100+20    & 68\% & 38\% & 27\% & 60\% \\
          & RSM100+50    & 78\% & 44\% & 32\% & 66\% \\
          & RSM100+100   & 81\% & 46\% & 36\% & 68\% \\
    \cmidrule(lr){1-6}
       \multirow{6}{*}{DP~\citep{chi2023diffusion}} 
          & RO           & 66\% & 49\% & 44\% & 66\% \\
          & SO           & 20\% & 10\% & 18\% & 28\% \\
          & RSM100+10    & 65\% & 53\% & 47\% & 67\% \\
          & RSM100+20    & 69\% & 58\% & 53\% & 70\% \\
          & RSM100+50    & 73\% & 62\% & 56\% & 73\% \\
          & RSM100+100   & 79\% & 66\% & 59\% & 78\% \\
   \bottomrule
    \end{tabular}
  }
\end{center}
\end{table}

Tab.~\ref{tab:il_results} summarizes success rates for ACT and DP under three dataset settings (RO, SO, RSM).
We highlight three findings:
(1) \textbf{Simulation-trained models achieve zero-shot success in the real world} (e.g., ACT 39\% on PullDrawer), reflecting ArtVIP’s high-fidelity visuals and physics that reduce the sim-to-real gap.
(2) With equal data volume, \textbf{real-world training outperforms simulation} (e.g., DP 49\% vs. 10\% on OpenCabinet), underscoring persistent sim-to-real challenges.
(3) \textbf{Mixing real and simulated data boosts performance} (e.g., SlideShelf: DP from 44\% to 59\%), indicating that articulated assets in ArtVIP align well with real-world data distributions.

\begin{figure}[!htbp]
  \centering
  \includegraphics[width=0.95\linewidth]{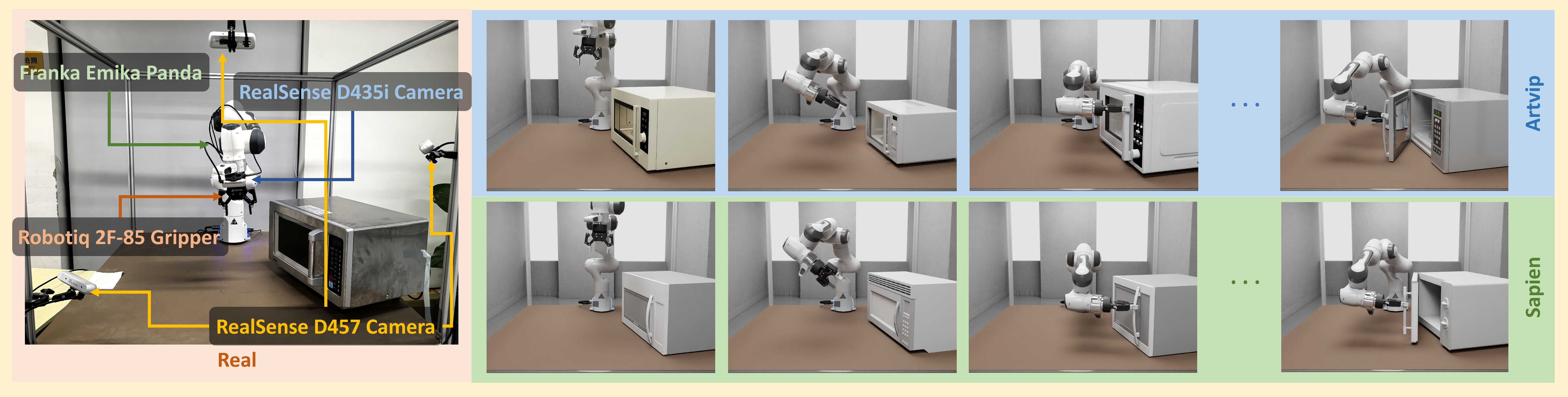}
  \caption{We collect data on five microwaves each from \ours and PartNet-Mobility.}
  \label{fig:micro_il}
\end{figure}

\subsection{Comparison with Other Assets via Imitation Learning.}

To validate the quality of \ours assets against other datasets, we conduct a digital-cousin comparison with PartNet-Mobility (Fig.~\ref{fig:micro_il}).
We select five microwave ovens from PartNet-Mobility and five from \ours. We select microwaves with pull-to-open doors and deliberately exclude button-triggered opening, as it is operationally trivial. We collect data via teleoperation following the same procedure as in the digital-twin experiments mentioned before, obtaining 100 simulated trajectories per microwave (500 in total). For the real-world task, we purchase an unseen microwave oven for which neither PartNet-Mobility nor \ours provides a corresponding digital-twin model.
We train ACT and DP under the following dataset settings: \textbf{(1) Real-Only (RO)}: 100 real-world trajectories; \textbf{(2) Sim-Only (SO)}: 500 simulated trajectories; \textbf{(3) Real–Sim–Mixed (RSM100+500)}: 100 real-world + 500 simulated trajectories. All runs use the same training hyperparameters.

Tab.~\ref{tab:il_cousin_results} summarizes success rates across three dataset settings. We highlight: \textbf{higher-quality \ours assets yield stronger zero-shot sim-to-real transfer under SO and higher success under RSM}, supporting the conclusion that higher-quality assets reduce the sim-to-real gap and lead to higher success rates.

\begin{table}[t]
\caption{Success rates of ACT and DP across dataset settings: RO (real-only), SO (sim-only), and RSM100+500, comparing \ours and PartNet-Mobility on the microwave door-pull task.}
\label{tab:il_cousin_results}
\begin{center}
\small
\begin{tabular}{llcc}
\toprule
Method & Dataset & \ours & PartNet-Mobility \\
\midrule
\multirow{3}{*}{ACT}
  & RO           & \multicolumn{2}{c}{56\%} \\
  & SO           & 41\% & 32\% \\
  & RSM100+500   & 79\% & 68\% \\
\cmidrule(lr){1-4}
\multirow{3}{*}{DP}
  & RO           & \multicolumn{2}{c}{62\%} \\
  & SO           & 45\% & 35\% \\
  & RSM100+500   & 83\% & 70\% \\
\bottomrule
\end{tabular}
\end{center}
\end{table}

\subsection{Reinforcement Learning in High-Fidelity Simulators}

Reinforcement learning (RL) requires training environments that mirror real-world physical and
perceptual complexity. To evaluate \ours, we design a CloseTrashcan task with a Franka arm and train a two-stage agent in Isaac Sim using EAGLE~\citep{zhao2025efficient} (Appendix Sec.~\ref{subsec:RL_APP}). We first train a PPO expert~\citep{schulman2017proximal} with low-level state inputs, then distill it into a visuomotor policy, applying EAGLE’s self-supervised attention masks and control-aware augmentation. We additionally use RandomConv~\citep{lee2019network} and Visual Matching~\citep{li2024evaluatingrealworldrobotmanipulation} to reduce the background gap between simulation and the real world.


\begin{table}[h]
\centering
\small
\caption{EAGLE vs. vision-based PPO: success rate across training checkpoints (k).}
\label{tab:rl_sim_vs_real}
\begin{tabular}{lccccc}
  \toprule
  \multirow{2}{*}{\textbf{Method}} & \multicolumn{5}{c}{\textbf{Training Iterations (k)}} \\
  \cmidrule(lr){2-6}
   & 100 & 200 & 300 & 400 & 500 \\
  \midrule
  EAGLE & 0.23 & 0.28 & 0.73 & 0.85 & 0.98 \\
  Vision-based PPO & 0.16 & 0.19 & 0.21 & 0.22 & 0.24 \\
  \bottomrule
\end{tabular}
\end{table}

We train in simulation and deploy the same policy in the real world. Tab.~\ref{tab:rl_sim_vs_real} reports success rates (more details in Appendix Sec.~\ref{subsec:Pearson}) at five checkpoints (300k--500k iterations), evaluated with 100 simulation trials and 30 real-world trials under varied initial object poses. The Pearson correlation coefficient is 0.9886, indicating a strong linear relationship between simulated and real-world performance, and supporting the fidelity of \ours.

\section{Limitation and Conclusion}
We introduced \ours, a high-quality dataset of articulated objects for robotic manipulation, featuring visual realism, accurate physical properties, and modular interaction capabilities. We validated its quality via diverse evaluations and demonstrated effectiveness in both imitation learning and reinforcement learning. Scaling remains bottlenecked by intensive human labor for asset modeling; future work will explore generative methods to automate synthesis, reduce manual effort, and broaden object diversity.

\newpage
\bibliography{iclr2026_conference}

@article{chi2023diffusion,
  title={Diffusion policy: Visuomotor policy learning via action diffusion},
  author={Chi, Cheng and Xu, Zhenjia and Feng, Siyuan and Cousineau, Eric and Du, Yilun and Burchfiel, Benjamin and Tedrake, Russ and Song, Shuran},
  journal={The International Journal of Robotics Research},
  pages={02783649241273668},
  year={2023},
  publisher={SAGE Publications Sage UK: London, England}
}

@misc{coumans2016pybullet,
  title={Pybullet, a python module for physics simulation for games, robotics and machine learning},
  author={Coumans, Erwin and Bai, Yunfei},
  year={2016}
}

@misc{NVIDIAPhysX,
  author       = {NVIDIA Corporation},
  title        = {NVIDIA PhysX SDK},
  year         = {2025},
  url          = {https://developer.nvidia.com/physx-sdk},
  urldate      = {2025-5-01}, 
}

@inproceedings{tasora2016chrono,
  title={Chrono: An open source multi-physics dynamics engine},
  author={Tasora, Alessandro and Serban, Radu and Mazhar, Hammad and Pazouki, Arman and Melanz, Daniel and Fleischmann, Jonathan and Taylor, Michael and Sugiyama, Hiroyuki and Negrut, Dan},
  booktitle={High Performance Computing in Science and Engineering: Second International Conference, HPCSE 2015, Sol{\'a}{\v{n}}, Czech Republic, May 25-28, 2015, Revised Selected Papers 2},
  pages={19--49},
  year={2016},
}

@misc{smith2005open,
  title={Open dynamics engine},
  author={Smith, Russell and others},
  year={2005}
}

@misc{pyrender,
author = {Matthew Matl},
title = {Pyrender},
year = {2019},
publisher = {GitHub},
journal = {GitHub repository},
howpublished = {\url{https://github.com/mmatl/pyrender}}
}

@misc{chociej2019orrbopenairemote,
      title={ORRB -- OpenAI Remote Rendering Backend}, 
      author={Maciek Chociej and Peter Welinder and Lilian Weng},
      year={2019},
      eprint={1906.11633},
      archivePrefix={arXiv},
      primaryClass={cs.GR},
      url={https://arxiv.org/abs/1906.11633}, 
}

@misc{rojtberg2024ogre,
  author = "{Rojtberg, Pavel and Rogers, David and Streeting, Steve and others}",
  title = "OGRE scene-oriented, flexible 3D engine",
  year = "2001 -- 2024",
  howpublished = "\url{https://www.ogre3d.org/}",
}

@misc{unity,
  author = {Unity Technologies},
  title = {Unity},
  version = {6.1},
  year = {2025.05.14},
  url = {https://unity.com/},
  note = {Game development platform},
}

@misc{isaacsim,
  author = {{Nvidia}},
  title = {NVIDIA Isaac Sim},
  version = {4.5},
  year = {2025.05.14},
  url = {https://developer.nvidia.com/isaac/sim},
  note = {Isaac Sim},
}

@misc{unrealengine,
  author = {Epic Games},
  title = {Unreal Engine},
  url = {https://www.unrealengine.com},
  version = {5.3},
  date = {2025-05-15},
    year={2025},
}

@misc{pbr,
  author = {Nvidia},
  title = {Understanding Physically-Based Rendering},
  url = {https://docs.omniverse.nvidia.com/simready/latest/simready-asset-creation/material-best-practices.html},
  version = {4.5},
  date = {2025-05-15},
    year={2025},
}

@misc{rtx,
  author = {Nvidia},
  title = {Omniverse RTX Renderer},
  url = {https://docs.omniverse.nvidia.com/materials-and-rendering/latest/rtx-renderer.html},
  version = {4.5},
  date = {2025-05-15},
    year={2025},
}

@inproceedings{shridhar2020alfred,
  title={Alfred: A benchmark for interpreting grounded instructions for everyday tasks},
  author={Shridhar, Mohit and Thomason, Jesse and Gordon, Daniel and Bisk, Yonatan and Han, Winson and Mottaghi, Roozbeh and Zettlemoyer, Luke and Fox, Dieter},
  booktitle={Proceedings of the IEEE/CVF Conference on Computer Vision and Pattern Recognition},
  pages={10740--10749},
  year={2020}
}

@article{macenski2022robot,
  title={Robot operating system 2: Design, architecture, and uses in the wild},
  author={Macenski, Steven and Foote, Tully and Gerkey, Brian and Lalancette, Chris and Woodall, William},
  journal={Science Robotics},
  volume={7},
  pages={eabm6074},
  year={2022}
}

@inproceedings{quigley2009ros,
  title={ROS: an open-source Robot Operating System},
  author={Quigley, Morgan and Conley, Ken and Gerkey, Brian and Faust, Josh and Foote, Tully and Leibs, Jeremy and Wheeler, Rob and Ng, Andrew Y and others},
  booktitle={ICRA workshop on open source software},
  volume={3},
  pages={5},
  year={2009}
}

@misc{dong2025digitaltwincataloglargescale,
      title={Digital Twin Catalog: A Large-Scale Photorealistic 3D Object Digital Twin Dataset}, 
      author={Zhao Dong and Ka Chen and Zhaoyang Lv and Hong-Xing Yu and Yunzhi Zhang and others},
      year={2025},
      eprint={2504.08541},
      archivePrefix={arXiv},
      primaryClass={cs.GR}
}

@article{kuang2023stanford,
  title={Stanford-orb: a real-world 3d object inverse rendering benchmark},
  author={Kuang, Zhengfei and Zhang, Yunzhi and Yu, Hong-Xing and Agarwala, Samir and Wu, Elliott and Wu, Jiajun and others},
  journal={Advances in Neural Information Processing Systems},
  volume={36},
  pages={46938--46957},
  year={2023}
}

@inproceedings{wang2024sm,
  title={SM 3: Self-supervised Multi-task Modeling with Multi-view 2D Images for Articulated Objects},
  author={Wang, Haowen and Zhao, Zhen and Jin, Zhao and Che, Zhengping and Qiao, Liang and Huang, Yakun and Fan, Zhipeng and Qiao, Xiuquan and Tang, Jian},
  booktitle={International Conference on Robotics and Automation},
  pages={12492--12498},
  year={2024}
}

@misc{xue2021omadobjectmodelarticulated,
      title={OMAD: Object Model with Articulated Deformations for Pose Estimation and Retrieval}, 
      author={Han Xue and Liu Liu and Wenqiang Xu and Haoyuan Fu and Cewu Lu},
      year={2021},
      eprint={2112.07334},
      archivePrefix={arXiv},
      primaryClass={cs.CV}
}

@article{chang2015shapenet,
  title={Shapenet: An information-rich 3d model repository},
  author={Chang, Angel X and Funkhouser, Thomas and Guibas, Leonidas and Hanrahan, Pat and Huang, Qixing and Li, Zimo and Savarese, Silvio and Savva, Manolis and Song, Shuran and Su, Hao and others},
  journal={arXiv preprint arXiv:1512.03012},
  year={2015}
}

@inproceedings{deitke2023objaverse,
  title={Objaverse: A universe of annotated 3d objects},
  author={Deitke, Matt and Schwenk, Dustin and Salvador, Jordi and Weihs, Luca and Michel, Oscar and VanderBilt, Eli and Schmidt, Ludwig and Ehsani, Kiana and Kembhavi, Aniruddha and Farhadi, Ali},
  booktitle={Proceedings of the IEEE/CVF Conference on Computer Vision and Pattern Recognition},
  pages={13142--13153},
  year={2023}
}

@inproceedings{koenig2004design,
  title={Design and use paradigms for gazebo, an open-source multi-robot simulator},
  author={Koenig, Nathan and Howard, Andrew},
  booktitle={IEEE/RSJ International Conference on Intelligent Robots and Systems},
  volume={3},
  pages={2149--2154},
  year={2004}
}

@inproceedings{todorov2012mujoco,
  title={MuJoCo: A physics engine for model-based control},
  author={Todorov, Emanuel and Erez, Tom and Tassa, Yuval},
  booktitle={IEEE/RSJ International Conference on Intelligent Robots and Systems},
  pages={5026--5033},
  year={2012}
}

@misc{Webots,
  AUTHOR = {Webots},
  TITLE  = {http://www.cyberbotics.com},
  NOTE   = {Open-source Mobile Robot Simulation Software},
  EDITOR = {Cyberbotics Ltd.},
  URL    = {http://www.cyberbotics.com},
  YEAR   = {2018}
}

@InProceedings{Xiang_2020_SAPIEN,
author = {Xiang, Fanbo and Qin, Yuzhe and Mo, Kaichun and Xia, Yikuan and Zhu, Hao and Liu, Fangchen and Liu, Minghua and Jiang, Hanxiao and Yuan, Yifu and Wang, He and Yi, Li and Chang, Angel X. and Guibas, Leonidas J. and Su, Hao},
title = {{SAPIEN}: A SimulAted Part-based Interactive ENvironment},
booktitle = {Proceedings of the IEEE/CVF Conference on Computer Vision and Pattern Recognition},
pages={11097--11107},
year={2020}
}

@inproceedings{robocasa2024,
  title={RoboCasa: Large-Scale Simulation of Everyday Tasks for Generalist Robots},
  author={Soroush Nasiriany and Abhiram Maddukuri and Lance Zhang and Adeet Parikh and Aaron Lo and Abhishek Joshi and Ajay Mandlekar and Yuke Zhu},
  booktitle={Robotics: Science and Systems},
  year={2024}
}

@inproceedings{dai2024acdc,
  title={Automated Creation of Digital Cousins for Robust Policy Learning},
  author={Tianyuan Dai and Josiah Wong and Yunfan Jiang and Chen Wang and Cem Gokmen and Ruohan Zhang and Jiajun Wu and Li Fei-Fei},
  booktitle={8th Annual Conference on Robot Learning},
  year={2024}
}

@misc{li2024behavior1khumancenteredembodiedai,
      title={BEHAVIOR-1K: A Human-Centered, Embodied AI Benchmark with 1,000 Everyday Activities and Realistic Simulation}, 
      author={Li, Chengshu and Zhang, Ruohan and Wong, Josiah and Gokmen, Cem and Srivastava, Sanjana and Mart{\'\i}n-Mart{\'\i}n, Roberto and Wang, Chen and Levine, Gabrael and Lingelbach, Michael and Sun, Jiankai and others},
      year={2024},
      eprint={2403.09227},
      archivePrefix={arXiv},
      primaryClass={cs.RO},
      url={https://arxiv.org/abs/2403.09227}, 
}

@article{chen2024urdformer,
  title={URDFormer: A Pipeline for Constructing Articulated Simulation Environments from Real-World Images},
  author={Zoey Chen and Aaron Walsman and Marius Memmel and Kaichun Mo and Alex Fang and Karthikeya Vemuri and Alan Wu and Dieter Fox and Abhishek Gupta},
  journal={arXiv preprint arXiv:2405.11656},
  year={2024}
}

@article{Eppner2024, 
    title={scene\_synthesizer: A Python Library for Procedural Scene Generation in Robot Manipulation}, 
    author={Clemens Eppner and Adithyavairavan Murali and Caelan Garrett and Rowland O'Flaherty and Tucker Hermans and Wei Yang and Dieter Fox},
    journal={Journal of Open Source Software},
    year={2024}
}

@misc{liu2023fewshotphysicallyawarearticulatedmesh,
      title={Few-Shot Physically-Aware Articulated Mesh Generation via Hierarchical Deformation}, 
      author={Xueyi Liu and Bin Wang and He Wang and Li Yi},
      year={2023},
      eprint={2308.10898},
      archivePrefix={arXiv},
      primaryClass={cs.CV}
}

@misc{li2020categorylevelarticulatedobjectpose,
      title={Category-Level Articulated Object Pose Estimation}, 
      author={Xiaolong Li and He Wang and Li Yi and Leonidas Guibas and A. Lynn Abbott and Shuran Song},
      year={2020},
      eprint={1912.11913},
      archivePrefix={arXiv},
      primaryClass={cs.CV},
      url={https://arxiv.org/abs/1912.11913}, 
}

@misc{liu2023selfsupervisedcategorylevelarticulatedobject,
      title={Self-Supervised Category-Level Articulated Object Pose Estimation with Part-Level SE(3) Equivariance}, 
      author={Xueyi Liu and Ji Zhang and Ruizhen Hu and Haibin Huang and He Wang and Li Yi},
      year={2023},
      eprint={2302.14268},
      archivePrefix={arXiv},
      primaryClass={cs.CV},
      url={https://arxiv.org/abs/2302.14268}, 
}

@misc{sun2023opdmultiopenabledetectionmultiple,
      title={OPDMulti: Openable Part Detection for Multiple Objects}, 
      author={Xiaohao Sun and Hanxiao Jiang and Manolis Savva and Angel Xuan Chang},
      year={2023},
      eprint={2303.14087},
      archivePrefix={arXiv},
      primaryClass={cs.CV},
      url={https://arxiv.org/abs/2303.14087}, 
}

@misc{brohan2023rt1roboticstransformerrealworld,
      title={RT-1: Robotics Transformer for Real-World Control at Scale}, 
      author={Brohan, Anthony and Brown, Noah and Carbajal, Justice and Chebotar, Yevgen and Dabis, Joseph and Finn, Chelsea and Gopalakrishnan, Keerthana and Hausman, Karol and Herzog, Alex and Hsu, Jasmine and others},
      year={2023},
      eprint={2212.06817},
      archivePrefix={arXiv},
      primaryClass={cs.RO}
}

@misc{zhao2023learningfinegrainedbimanualmanipulation,
      title={Learning Fine-Grained Bimanual Manipulation with Low-Cost Hardware}, 
      author={Tony Z. Zhao and Vikash Kumar and Sergey Levine and Chelsea Finn},
      year={2023},
      eprint={2304.13705},
      archivePrefix={arXiv},
      primaryClass={cs.RO}
}

@misc{geminiroboticsteam2025geminiroboticsbringingai,
      title={Gemini Robotics: Bringing AI into the Physical World}, 
      author={Gemini Robotics Team},
      year={2025},
      eprint={2503.20020},
      archivePrefix={arXiv},
      primaryClass={cs.RO}
}

@misc{intelligence2025pi05visionlanguageactionmodelopenworld,
      title={$\pi_{0.5}$: a Vision-Language-Action Model with Open-World Generalization}, 
      author={Physical Intelligence},
      year={2025},
      eprint={2504.16054},
      archivePrefix={arXiv},
      primaryClass={cs.LG} 
}

@misc{brohan2023rt2visionlanguageactionmodelstransfer,
      title={RT-2: Vision-Language-Action Models Transfer Web Knowledge to Robotic Control}, 
      author={Brohan, Anthony and Brown, Noah and Carbajal, Justice and Chebotar, Yevgen and Chen, Xi and Choromanski, Krzysztof and Ding, Tianli and Driess, Danny and Dubey, Avinava and Finn, Chelsea and others},
      year={2023},
      eprint={2307.15818},
      archivePrefix={arXiv},
      primaryClass={cs.RO},
}

@misc{kim2024openvlaopensourcevisionlanguageactionmodel,
      title={OpenVLA: An Open-Source Vision-Language-Action Model}, 
      author={Moo Jin Kim and Karl Pertsch and Siddharth Karamcheti and Ted Xiao and Ashwin Balakrishna and others},
      year={2024},
      eprint={2406.09246},
      archivePrefix={arXiv},
      primaryClass={cs.RO},
}

@article{billard2019trends,
  title={Trends and challenges in robot manipulation},
  author={Billard, Aude and Kragic, Danica},
  journal={Science},
  volume={364},
  pages={eaat8414},
  year={2019},
}

@article{spong2020robot,
  title={Robot modeling and control},
  author={Spong, Mark W and Hutchinson, Seth and Vidyasagar, M},
  journal={John Wiley \&amp},
  year={2020}
}

@article{ramasubramanian2022digital,
  title={Digital twin for human--robot collaboration in manufacturing: Review and outlook},
  author={Ramasubramanian, Aswin K and Mathew, Robins and Kelly, Matthew and Hargaden, Vincent and Papakostas, Nikolaos},
  journal={Applied Sciences},
  volume={12},
  pages={4811},
  year={2022},
}

@article{makoviychuk2021isaac,
  title={Isaac gym: High performance gpu-based physics simulation for robot learning},
  author={Makoviychuk, Viktor and Wawrzyniak, Lukasz and Guo, Yunrong and Lu, Michelle and Storey, Kier and Macklin, Miles and Hoeller, David and Rudin, Nikita and Allshire, Arthur and Handa, Ankur and others},
  journal={arXiv preprint arXiv:2108.10470},
  year={2021}
}

@inproceedings{
zhang2023crossformer,
title={Crossformer: Transformer Utilizing Cross-Dimension Dependency for Multivariate Time Series Forecasting},
author={Yunhao Zhang and Junchi Yan},
booktitle={International Conference on Learning Representations},
year={2023},
}

@misc{wang2024scalingproprioceptivevisuallearningheterogeneous,
      title={Scaling Proprioceptive-Visual Learning with Heterogeneous Pre-trained Transformers}, 
      author={Lirui Wang and Xinlei Chen and Jialiang Zhao and Kaiming He},
      year={2024},
      eprint={2409.20537},
      archivePrefix={arXiv},
      primaryClass={cs.RO}
}

@InProceedings{ge2024behavior,
    title={BEHAVIOR Vision Suite: Customizable Dataset Generation via Simulation},
    author={Ge, Yunhao and Tang, Yihe and Xu, Jiashu and Gokmen, Cem and Li, Chengshu and Ai, Wensi and Martinez, Benjamin Jose and Aydin, Arman and Anvari, Mona and Chakravarthy, Ayush K and others},
    booktitle = {Proceedings of the IEEE/CVF Conference on Computer Vision and Pattern Recognition},
    year={2024},
    pages={22401-22412}
}

@article{choi2021use,
  title={On the use of simulation in robotics: Opportunities, challenges, and suggestions for moving forward},
  author={Choi, HeeSun and Crump, Cindy and Duriez, Christian and Elmquist, Asher and Hager, Gregory and Han, David and Hearl, Frank and Hodgins, Jessica and Jain, Abhinandan and Leve, Frederick and others},
  journal={Proceedings of the National Academy of Sciences},
  volume={118},
  pages={e1907856118},
  year={2021},
}

@article{straub2019replica,
  title={The replica dataset: A digital replica of indoor spaces},
  author={Straub, Julian and Whelan, Thomas and Ma, Lingni and Chen, Yufan and Wijmans, Erik and Green, Simon and Engel, Jakob J and Mur-Artal, Raul and Ren, Carl and Verma, Shobhit and others},
  journal={arXiv preprint arXiv:1906.05797},
  year={2019}
}

@inproceedings{ramakrishnan2021hm3d,
  title={Habitat-Matterport 3D Dataset ({HM}3D): 1000 Large-scale 3D Environments for Embodied {AI}},
  author={Santhosh Kumar Ramakrishnan and Aaron Gokaslan and Erik Wijmans and Oleksandr Maksymets and Alexander Clegg and John M Turner and Eric Undersander and Wojciech Galuba and Andrew Westbury and Angel X Chang and Manolis Savva and Yili Zhao and Dhruv Batra},
  booktitle={Thirty-fifth Conference on Neural Information Processing Systems Datasets and Benchmarks Track},
  year={2021},
  url={https://arxiv.org/abs/2109.08238}
}

@article{torne2024rialto,
        author    = {Torne, Marcel 
                    and Simeonov, Anthony 
                    and Li, Zechu 
                    and Chan, April 
                    and Chen, Tao 
                    and Gupta, Abhishek 
                    and Agrawal, Pulkit},
        title     = {Reconciling Reality Through Simulation: A Real-to-Sim-to-Real Approach for Robust Manipulation},
        journal   = {Arxiv},
        year      = {2024},
      }

@article{do2025watch,
      title={Watch Less, Feel More: Sim-to-Real RL for Generalizable Articulated Object Manipulation via Motion Adaptation and Impedance Control},
      author={Do, Tan-Dzung and Gireesh, Nandiraju and Wang, Jilong and Wang, He},
      journal={arXiv preprint arXiv:2502.14457},
      year={2025}
    }

@article{wu2024robomind,
  title={Robomind: Benchmark on multi-embodiment intelligence normative data for robot manipulation},
  author={Wu, Kun and Hou, Chengkai and Liu, Jiaming and Che, Zhengping and Ju, Xiaozhu and Yang, Zhuqin and Li, Meng and Zhao, Yinuo and Xu, Zhiyuan and Yang, Guang and others},
  journal={arXiv preprint arXiv:2412.13877},
  year={2024}
}

@article{liu2024singapo,
  title={SINGAPO: Single Image Controlled Generation of Articulated Parts in Objects},
  author={Liu, Jiayi and Iliash, Denys and Chang, Angel X and Savva, Manolis and Mahdavi-Amiri, Ali},
  journal={arXiv preprint arXiv:2410.16499},
  year={2024}
}

@article{ai2thor,
  author={Eric Kolve and Roozbeh Mottaghi and Winson Han and
          Eli VanderBilt and Luca Weihs and Alvaro Herrasti and
          Daniel Gordon and Yuke Zhu and Abhinav Gupta and
          Ali Farhadi},
  title={{AI2-THOR: An Interactive 3D Environment for Visual AI}},
  journal={arXiv},
  year={2017}
}

@misc{puig2023habitat30cohabitathumans,
      title={Habitat 3.0: A Co-Habitat for Humans, Avatars and Robots}, 
      author={Xavier Puig and Eric Undersander and Andrew Szot and Mikael Dallaire Cote and Tsung-Yen Yang and Ruslan Partsey and Ruta Desai and Alexander William Clegg and Michal Hlavac and So Yeon Min and others},
      year={2023},
      eprint={2310.13724},
      archivePrefix={arXiv},
      primaryClass={cs.HC}
}

@article{szot2021habitat,
  title={Habitat 2.0: Training home assistants to rearrange their habitat},
  author={Szot, Andrew and Clegg, Alexander and Undersander, Eric and Wijmans, Erik and Zhao, Yili and Turner, John and Maestre, Noah and Mukadam, Mustafa and Chaplot, Devendra Singh and Maksymets, Oleksandr and others},
  journal={Advances in Neural Information Processing Systems},
  volume={34},
  pages={251--266},
  year={2021}
}

@inproceedings{habitat19iccv,
  title={Habitat: A platform for embodied ai research},
  author={Savva, Manolis and Kadian, Abhishek and Maksymets, Oleksandr and Zhao, Yili and Wijmans, Erik and Jain, Bhavana and Straub, Julian and Liu, Jia and Koltun, Vladlen and Malik, Jitendra and others},
  booktitle={Proceedings of the IEEE/CVF international conference on computer vision},
  pages={9339--9347},
  year={2019}
}

@inproceedings{li2022igibson,
  title={iGibson 2.0: Object-Centric Simulation for Robot Learning of Everyday Household Tasks},
  author={Li, Chengshu and Xia, Fei and Mart{\'\i}n-Mart{\'\i}n, Roberto and Lingelbach, Michael and Srivastava, Sanjana and Shen, Bokui and Vainio, Kent Elliott and Gokmen, Cem and Dharan, Gokul and Jain, Tanish and others},
  booktitle={5th Annual Conference on Robot Learning},
  year={2022}
}

@inproceedings{shen2021igibson,
  title={igibson 1.0: A simulation environment for interactive tasks in large realistic scenes},
  author={Shen, Bokui and Xia, Fei and Li, Chengshu and Mart{\'\i}n-Mart{\'\i}n, Roberto and Fan, Linxi and Wang, Guanzhi and P{\'e}rez-D’Arpino, Claudia and Buch, Shyamal and Srivastava, Sanjana and Tchapmi, Lyne and others},
  booktitle={IEEE/RSJ International Conference on Intelligent Robots and Systems},
  pages={7520--7527},
  year={2021},
}

@article{su2024artformer,
  title={Artformer: Controllable generation of diverse 3d articulated objects},
  author={Su, Jiayi and Feng, Youhe and Li, Zheng and Song, Jinhua and He, Yangfan and Ren, Botao and Xu, Botian},
  journal={arXiv preprint arXiv:2412.07237},
  year={2024}
}

@article{liu2023meshdiffusion,
  title={Meshdiffusion: Score-based generative 3d mesh modeling},
  author={Liu, Zhen and Feng, Yao and Black, Michael J and Nowrouzezahrai, Derek and Paull, Liam and Liu, Weiyang},
  journal={arXiv preprint arXiv:2303.08133},
  year={2023}
}

@article{long2023wonder3d,
  title={Wonder3D: Single Image to 3D using Cross-Domain Diffusion},
  author={Long, Xiaoxiao and Guo, Yuan-Chen and Lin, Cheng and Liu, Yuan and Dou, Zhiyang and Liu, Lingjie and Ma, Yuexin and Zhang, Song-Hai and Habermann, Marc and Theobalt, Christian and others},
  journal={arXiv preprint arXiv:2310.15008},
  year={2023}
}

@misc{nesti2025simprivesimulationframeworkphysical,
      title={SimPRIVE: a Simulation framework for Physical Robot Interaction with Virtual Environments}, 
      author={Federico Nesti and Gianluca D'Amico and Mauro Marinoni and Giorgio Buttazzo},
      year={2025},
      eprint={2504.21454},
      archivePrefix={arXiv},
      primaryClass={cs.RO},
}

@misc{han2025re3simgeneratinghighfidelitysimulation,
      title={Re$^3$Sim: Generating High-Fidelity Simulation Data via 3D-Photorealistic Real-to-Sim for Robotic Manipulation}, 
      author={Xiaoshen Han and Minghuan Liu and Yilun Chen and Junqiu Yu and Xiaoyang Lyu and Yang Tian and Bolun Wang and Weinan Zhang and Jiangmiao Pang},
      year={2025},
      eprint={2502.08645},
      archivePrefix={arXiv},
      primaryClass={cs.RO},
}

@misc{embleyriches2025unrealroboticslabhighfidelity,
      title={Unreal Robotics Lab: A High-Fidelity Robotics Simulator with Advanced Physics and Rendering}, 
      author={Jonathan Embley-Riches and Jianwei Liu and Simon Julier and Dimitrios Kanoulas},
      year={2025},
      eprint={2504.14135},
      archivePrefix={arXiv},
      primaryClass={cs.RO},
}

@inproceedings{wang2025vggt,
  title={VGGT: Visual Geometry Grounded Transformer},
  author={Wang, Jianyuan and Chen, Minghao and Karaev, Nikita and Vedaldi, Andrea and Rupprecht, Christian and Novotny, David},
  booktitle={Proceedings of the IEEE/CVF Conference on Computer Vision and Pattern Recognition},
  year={2025}
}

@article{van2008visualizing,
  title={Visualizing data using t-SNE.},
  author={Van der Maaten, Laurens and Hinton, Geoffrey},
  journal={Journal of machine learning research},
  volume={9},
  number={11},
  year={2008}
}

@inproceedings{radford2021learning,
  title={Learning transferable visual models from natural language supervision},
  author={Radford, Alec and Kim, Jong Wook and Hallacy, Chris and Ramesh, Aditya and Goh, Gabriel and Agarwal, Sandhini and Sastry, Girish and Askell, Amanda and Mishkin, Pamela and Clark, Jack and others},
  booktitle={International Conference on Machine Learning},
  pages={8748--8763},
  year={2021},
}

@inproceedings{zhao2025efficient,
  author    = {Zhao, Yinuo and Wu, Kun and Yi, Tianjiao and Xu, Zhiyuan and Che, Zhengping  and
               Liu, Chi Harold and Tang, Jian},
  title     = {Efficient Training of Generalizable Visuomotor Policies via Control-Aware Augmentation},
  booktitle = {Proceedings of the 24th International Conference on Autonomous Agents and Multiagent Systems},
  year      = {2025},
}

@article{schulman2017proximal,
  title={Proximal policy optimization algorithms},
  author={Schulman, John and Wolski, Filip and Dhariwal, Prafulla and Radford, Alec and Klimov, Oleg},
  journal={arXiv preprint arXiv:1707.06347},
  year={2017}
}

@article{lee2019network,
  title={Network randomization: A simple technique for generalization in deep reinforcement learning},
  author={Lee, Kimin and Lee, Kibok and Shin, Jinwoo and Lee, Honglak},
  journal={International Conference on Learning Representations},
  year={2019}
}

@misc{IsaacSimJointTuning,
  author = {NVIDIA},
  title = {Joint Tuning --- {Isaac} {Sim} Documentation},
  year = {2025},
  url = {https://docs.isaacsim.omniverse.nvidia.com/latest/robot_setup/joint_tuning.html#gain-tuning},
  urldate = {2025-5-14}, 
}

@misc{yang2022dsgnetlearningdisentangledstructure,
      title={DSG-Net: Learning Disentangled Structure and Geometry for 3D Shape Generation}, 
      author={Jie Yang and Kaichun Mo and Yu-Kun Lai and Leonidas J. Guibas and Lin Gao},
      year={2022},
      eprint={2008.05440},
      archivePrefix={arXiv},
      primaryClass={cs.GR},
      url={https://arxiv.org/abs/2008.05440}, 
}

@misc{koo2024saladpartlevellatentdiffusion,
      title={SALAD: Part-Level Latent Diffusion for 3D Shape Generation and Manipulation}, 
      author={Juil Koo and Seungwoo Yoo and Minh Hieu Nguyen and Minhyuk Sung},
      year={2024},
      eprint={2303.12236},
      archivePrefix={arXiv},
      primaryClass={cs.CV},
      url={https://arxiv.org/abs/2303.12236}, 
}

@misc{xu2023dmv3ddenoisingmultiviewdiffusion,
      title={DMV3D: Denoising Multi-View Diffusion using 3D Large Reconstruction Model}, 
      author={Yinghao Xu and Hao Tan and Fujun Luan and Sai Bi and Peng Wang and Jiahao Li and Zifan Shi and Kalyan Sunkavalli and Gordon Wetzstein and Zexiang Xu and Kai Zhang},
      year={2023},
      eprint={2311.09217},
      archivePrefix={arXiv},
      primaryClass={cs.CV},
      url={https://arxiv.org/abs/2311.09217}, 
}

@misc{3dwarehouse,
  author = {Trimble Inc.},
  title = {3D Warehouse},
  url = {https://3dwarehouse.sketchup.com/},
  note = {Online 3D model repository for SketchUp},
  year = {2024},
  urldate = {2024-12-19}
}

@misc{turbosquid,
  author = {TurboSquid Inc.},
  title = {TurboSquid},
  url = {https://www.turbosquid.com/},
  note = {Professional 3D model marketplace},
  year = {2024},
  urldate = {2024-12-19}
}

@misc{liu2024cagecontrollablearticulationgeneration,
      title={CAGE: Controllable Articulation GEneration}, 
      author={Jiayi Liu and Hou In Ivan Tam and Ali Mahdavi-Amiri and Manolis Savva},
      year={2024},
      eprint={2312.09570},
      archivePrefix={arXiv},
      primaryClass={cs.CV},
      url={https://arxiv.org/abs/2312.09570}, 
}

@misc{qiu2025articulateanymeshopenvocabulary3d,
      title={Articulate AnyMesh: Open-Vocabulary 3D Articulated Objects Modeling}, 
      author={Xiaowen Qiu and Jincheng Yang and Yian Wang and Zhehuan Chen and Yufei Wang and Tsun-Hsuan Wang and Zhou Xian and Chuang Gan},
      year={2025},
      eprint={2502.02590},
      archivePrefix={arXiv},
      primaryClass={cs.CV},
      url={https://arxiv.org/abs/2502.02590}, 
}

@misc{mandi2024real2codereconstructarticulatedobjects,
      title={Real2Code: Reconstruct Articulated Objects via Code Generation}, 
      author={Zhao Mandi and Yijia Weng and Dominik Bauer and Shuran Song},
      year={2024},
      eprint={2406.08474},
      archivePrefix={arXiv},
      primaryClass={cs.CV},
      url={https://arxiv.org/abs/2406.08474}, 
}

@misc{le2025articulateanythingautomaticmodelingarticulated,
      title={Articulate-Anything: Automatic Modeling of Articulated Objects via a Vision-Language Foundation Model}, 
      author={Long Le and Jason Xie and William Liang and Hung-Ju Wang and Yue Yang and Yecheng Jason Ma and Kyle Vedder and Arjun Krishna and Dinesh Jayaraman and Eric Eaton},
      year={2025},
      eprint={2410.13882},
      archivePrefix={arXiv},
      primaryClass={cs.CV},
      url={https://arxiv.org/abs/2410.13882}, 
}

@misc{li2024evaluatingrealworldrobotmanipulation,
      title={Evaluating Real-World Robot Manipulation Policies in Simulation}, 
      author={Xuanlin Li and Kyle Hsu and Jiayuan Gu and Karl Pertsch and Oier Mees and Homer Rich Walke and Chuyuan Fu and Ishikaa Lunawat and Isabel Sieh and Sean Kirmani and Sergey Levine and Jiajun Wu and Chelsea Finn and Hao Su and Quan Vuong and Ted Xiao},
      year={2024},
      eprint={2405.05941},
      archivePrefix={arXiv},
      primaryClass={cs.RO},
      url={https://arxiv.org/abs/2405.05941}, 
}

@article{lin2025splart,
  title={SplArt: Articulation Estimation and Part-Level Reconstruction with 3D Gaussian Splatting},
  author={Lin, Shengjie and Fang, Jiading and Irshad, Muhammad Zubair and Guizilini, Vitor Campagnolo and Ambrus, Rares Andrei and Shakhnarovich, Greg and Walter, Matthew R},
  journal={arXiv preprint arXiv:2506.03594},
  year={2025}
}

@misc{liu2025artgsbuildinginteractablereplicas,
      title={ArtGS: Building Interactable Replicas of Complex Articulated Objects via Gaussian Splatting}, 
      author={Yu Liu and Baoxiong Jia and Ruijie Lu and Junfeng Ni and Song-Chun Zhu and Siyuan Huang},
      year={2025},
      eprint={2502.19459},
      archivePrefix={arXiv},
      primaryClass={cs.CV},
      url={https://arxiv.org/abs/2502.19459}, 
}

@misc{guo2025articulatedgs,
  title={ArticulatedGS: Self-supervised Digital Twin Modeling of Articulated Objects Using 3D Gaussian Splatting},
  author={Guo, Junfu and others},
  year={2025}
}
\bibliographystyle{iclr2026_conference}

\newpage
\appendix
\section{Articulated Objects}
\label{subsec:Articulated_Objects}
\ours comprises \totalNum articulated objects, encompassing 9 categories and \totalCat subcategories, with a total of 2156 prismatic joints and 1809 revolute joints. The detailed breakdown, including approximate human labor time, is presented in Tab.~\ref{tab:appendix_object_details}.

\begin{table}[htbp]
\centering
\caption{Detailed breakdown of object categories, modeling time (each), physics tuning time (each), and count.}
\label{tab:appendix_object_details}
\begin{tabular}{lllll}
\toprule
 category & subcategories & modeling time & physics tuning time & count \\
\midrule
\multirow{6}{*}{furniture} & chair & 2h & 0.3h & 23 \\
\cmidrule(lr){2-5}
 & table & 1.5h & 0.2h & 131 \\
\cmidrule(lr){2-5}
 & cabinet & 3.1h & 0.4h & 183 \\
\cmidrule(lr){2-5}
 & cupboard & 15h & 2h & 11 \\
\cmidrule(lr){2-5}
 & bed & 3h & 0.3h & 28 \\
\cmidrule(lr){2-5}
 & home decor & 2.5h & 0.3h & 30 \\
\midrule
\multirow{1}{*}{kitchenware} & cookware & 1.9h & 0.3h & 81 \\
\midrule
\multirow{9}{*}{kitchen appliances} & coffee machine & 3h & 0.3h & 14 \\
\cmidrule(lr){2-5}
 & built-in oven & 5h & 0.3h & 14 \\
\cmidrule(lr){2-5}
 & microwave & 3h & 0.3h & 8 \\
\cmidrule(lr){2-5}
 & oven & 4h & 0.3h & 11 \\
\cmidrule(lr){2-5}
 & dishwasher & 5h & 0.4h & 19 \\
\cmidrule(lr){2-5}
 & water dispenser & 3h & 0.3h & 6 \\
\cmidrule(lr){2-5}
 & rice cooker & 3h & 0.3h & 14 \\
\cmidrule(lr){2-5}
 & fridge & 6h & 0.5h & 22 \\
\cmidrule(lr){2-5}
 & juicer & 4h & 0.3h & 6 \\
\midrule
\multirow{3}{*}{fixtures} & faucet & 2h & 0.2h & 14 \\
\cmidrule(lr){2-5}
 & toilet & 4h & 0.4h & 14 \\
\cmidrule(lr){2-5}
 & door & 2h & 0.3h & 10 \\
\midrule
\multirow{6}{*}{appliances} & computer & 2.5h & 0.3h & 13 \\
\cmidrule(lr){2-5}
 & fan & 1.8h & 0.3h & 34 \\
\cmidrule(lr){2-5}
 & air conditioner & 4h & 0.3h & 3 \\
\cmidrule(lr){2-5}
 & washing machine & 5.7h & 0.5h & 30 \\
\cmidrule(lr){2-5}
 & speaker & 1.5h & 0.3h & 14 \\
\cmidrule(lr){2-5}
 & floor lamp & 1h & 0.3h & 28 \\
\midrule
\multirow{3}{*}{cleaning tools} & mop & 2h & 0.3h & 8 \\
\cmidrule(lr){2-5}
 & pump bottle & 2h & 0.3h & 14 \\
\cmidrule(lr){2-5}
 & trash can & 2h & 0.3h & 18 \\
\midrule
\multirow{4}{*}{stationery} & scissors & 1h & 0.2h & 28 \\
\cmidrule(lr){2-5}
 & stapler & 1.5h & 0.2h & 11 \\
\cmidrule(lr){2-5}
 & utility knife & 1h & 0.2h & 19 \\
\cmidrule(lr){2-5}
 & folder & 0.5h & 0.2h & 8 \\
\midrule
\multirow{3}{*}{storage} & storage box & 2h & 0.3h & 25 \\
\cmidrule(lr){2-5}
 & toolbox & 2.5h & 0.3h & 22 \\
\cmidrule(lr){2-5}
 & cardboard box & 1.5h & 0.2h & 28 \\
\midrule
\multirow{2}{*}{Mechanical equipment} & electrical equipment & 3h & 0.3h & 17 \\
\cmidrule(lr){2-5}
 & non-electrical equipment & 3h & 0.3h & 33 \\
\bottomrule
\end{tabular}
\end{table}

\section{Scenes}
\label{subsec:Scenes}
\begin{figure}[!b]
  \centering
  \begin{subfigure}[b]{0.49\linewidth}
    \includegraphics[width=\linewidth]{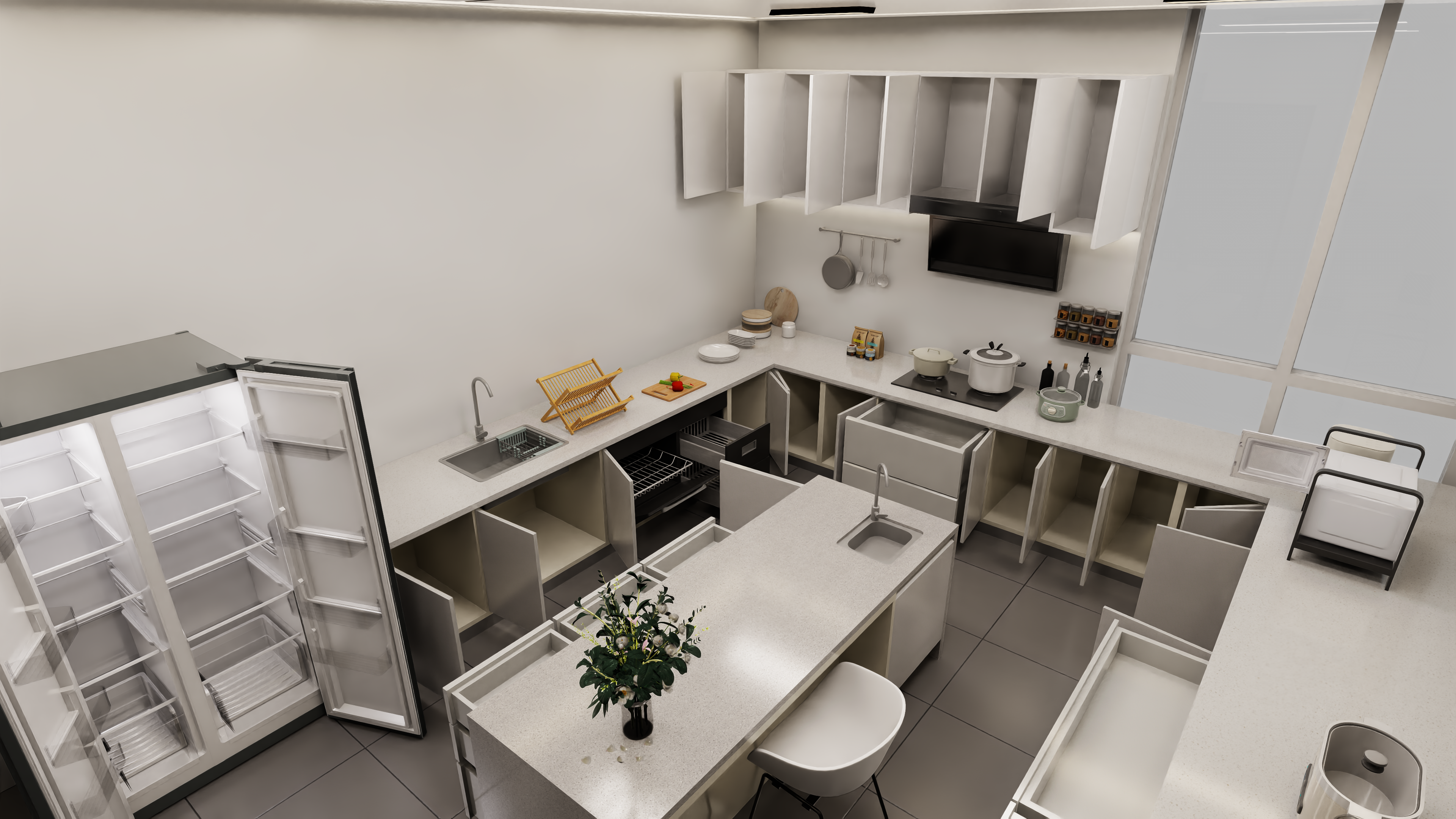}
    \caption{Kitchen}
  \end{subfigure}\hfill
  \begin{subfigure}[b]{0.49\linewidth}
    \includegraphics[width=\linewidth]{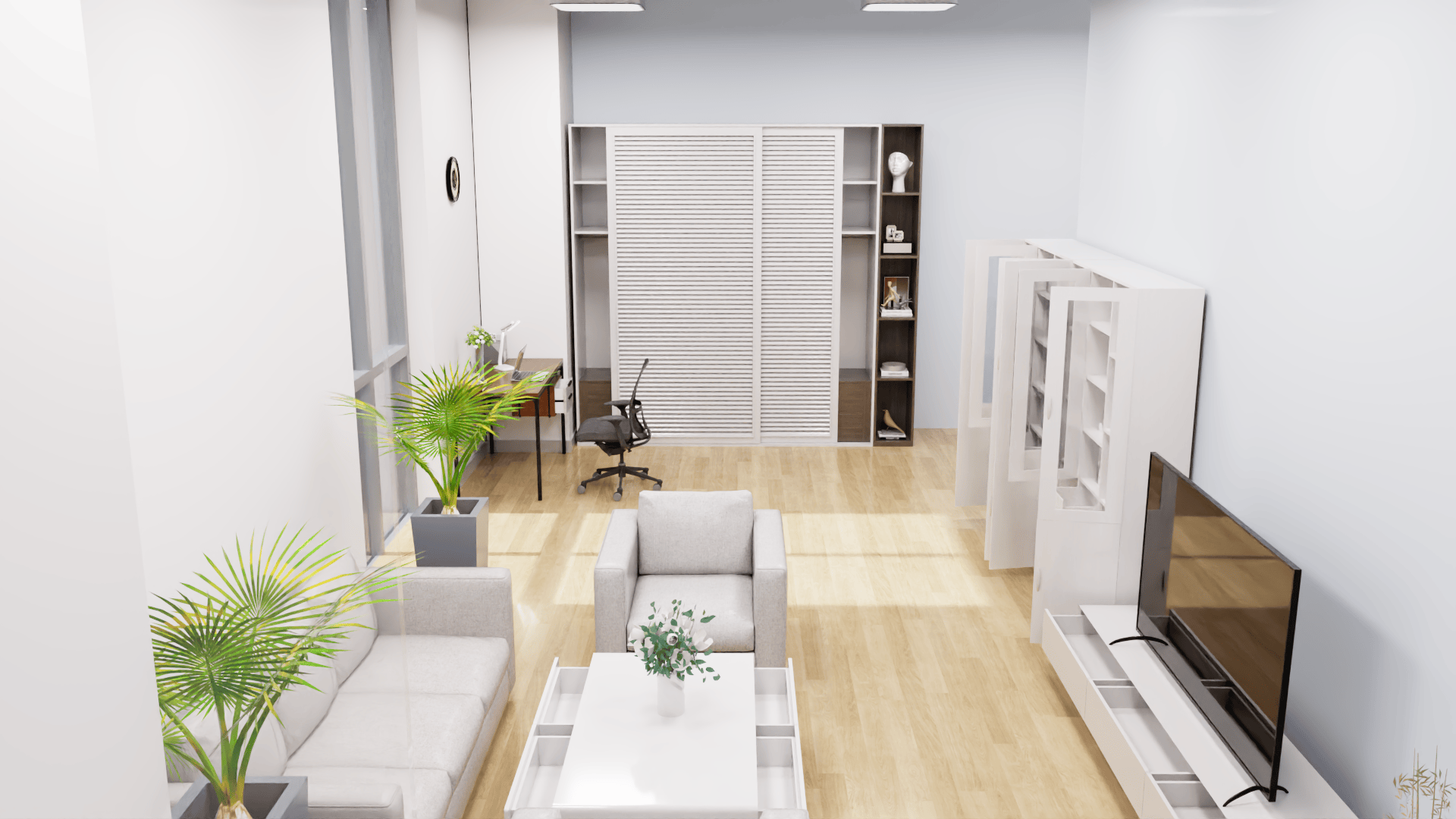}
    \caption{Small livingroom}
  \end{subfigure}
  \caption{Scenes: all articulated joints in the open state.}
  \label{fig:appendix_scene}
\end{figure}

We provide sim-ready complex, dynamic environments—six in total: childrenroom, diningroom, kitchen, kitchen with parlor, large livingroom, and small livingroom (see Fig.~\ref{fig:appendix_scene} for two example scenes). Every object in these environments, including fixed furniture, supports physical interaction. This includes switches, small appliances, plush toys, laptops, books, spice jars, and more. For example, the kitchen environment contains a total of 65 joints, and all objects can be used just like their real-world counterparts. Robots can operate the light switch on the wall, open the refrigerator door, place items on shelves, or challenge their motion capabilities by crouching to open drawers beneath the stove top.
Additionally, users can freely place the \totalNum articulated objects provided in ArtVIP into any of these environments via the Isaac Sim GUI, enabling the creation of rich robot interaction scenarios such as grasping, pulling, pressing, and placing. Users can also utilize open-source tools like mjcf2usd and urdf2usd to convert assets from other datasets into the USD format, allowing seamless integration with ArtVIP assets. This kind of sim-ready, complex environment is currently unique to ArtVIP. Moreover, the ability to edit and save assets directly through a GUI reflects an open-source spirit that is not yet common in other datasets.

\section{Annotations}
\label{subsec:annotations}
\begin{figure}[t!]
  \centering
  \includegraphics[width=0.85\linewidth]{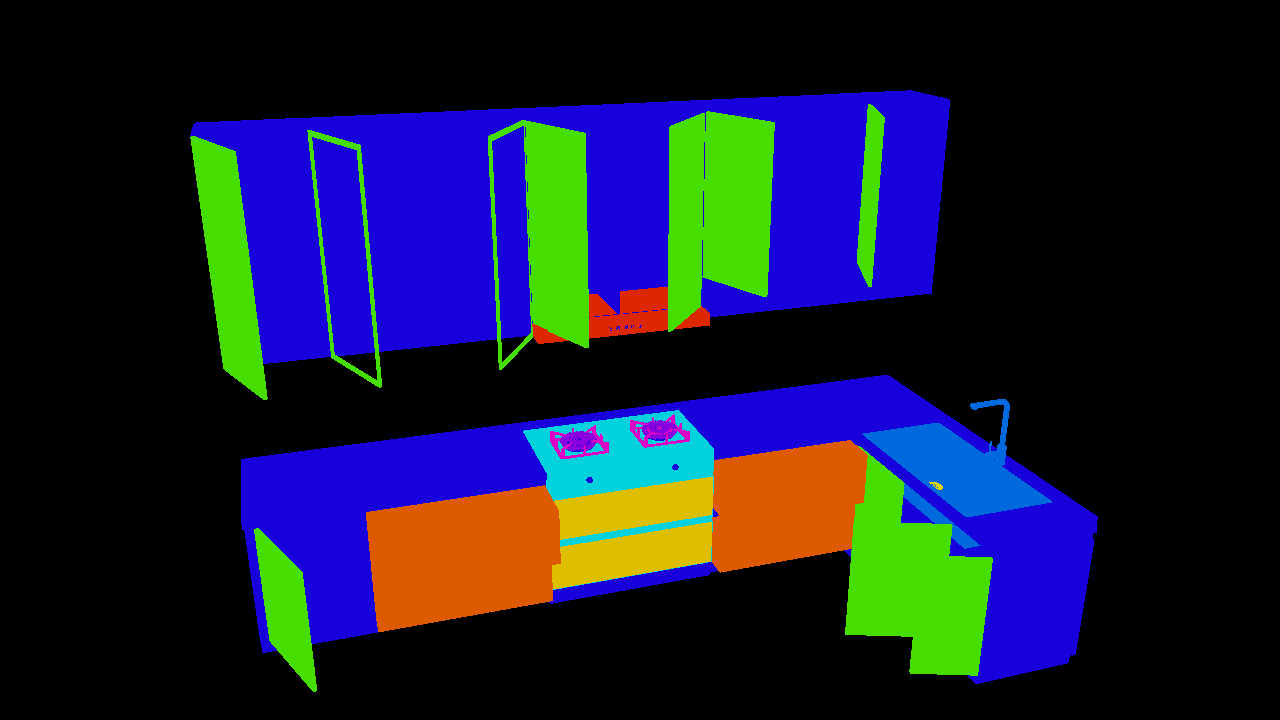}
  \caption{
     Segmentation result of the kitchen scene.
  }
  \label{fig:appendix_seg}
\end{figure}
Annotations in \ours provide objective descriptions of object parts, thereby supporting robots' ability to infer task-appropriate interaction behaviors. We further argue that annotations are most meaningful when aligned with consistent modeling standards. For example, for a desk, modelers often merge the legs and tabletop into a single mesh, which limits part-level annotation based on distinct interaction functions. To address this, we highlight functional components in Tab.~\ref{tab:annotation_labels} that frequently participate in interactions yet are commonly overlooked during mesh segmentation. An example segmentation is shown in Fig.~\ref{fig:appendix_seg}.

\begin{table}[t]
\centering
\caption{Annotation labels and descriptions in \ours.}
\label{tab:annotation_labels}
\begin{tabular}{lp{0.72\textwidth}}
\toprule
Label & Description \\
\midrule
armrest & Chair armrest \\
backrest & Chair backrest \\
ball\_handle & Handle for lifting the main body, such as the handle of a toolbox \\
blade & Blade of a utility knife, scissors, or fan blades \\
body & Parts that need labeling excluding base and lid \\
button & Applies to all push-button switch components of models \\
door & Door of cabinets, refrigerators, ovens, etc. \\
drawer & Drawer of cabinets, refrigerators, toolboxes, etc. \\
front\_cover & Cover of a folder \\
fun\_guard & Fan protective cover \\
handle & Any handles \\
headrest & Chair headrest \\
jaw & Head of pliers, the part that contacts the gripped item \\
keyboard & Computer keyboard \\
knob & Applies to all rotary switch components of models \\
lid & Such as cardboard box lid, electric steamer lid, trash can lid \\
light & All types of lights \\
mop\_head & Mop head \\
pedal & Foot pedal, such as on a step-on trash can \\
pipe & Water pipe part of faucet \\
plate & All types of plates \\
pole & Rod-shaped component \\
portafilter & A handle holds the coffee grounds \\
pot & Inner pot of rice cookers, steamers, etc. \\
rack & Rack in an oven, refrigerator door shelf \\
roller & Washing machine drum \\
screen & Electronic product screen \\
seat & Chair seat \\
shelf & Shelf part of cabinets, refrigerators, etc. \\
spout & Spout of a pump bottle, water dispenser, etc. \\
stapler\_magazine & Staple compartment of a stapler \\
tabletop & Top surface of a table \\
toilet\_seat & Toilet seat \\
touch\_pad & Computer touchpad \\
wheel & Chair wheels \\
\bottomrule
\end{tabular}
\end{table}

\section{Modeling Standards}
\label{subsec:Modeling_Standards}
In simulation systems, the use of high-quality meshes, textures, and materials confers several advantages. High-fidelity visuals reduce the disparity between simulation and reality~\citep{nesti2025simprivesimulationframeworkphysical}, thereby narrowing the sim-to-real gap and enabling robotic policies to be deployed in real-world environments with minimal or even zero-shot adaptation~\citep{han2025re3simgeneratinghighfidelitysimulation,embleyriches2025unrealroboticslabhighfidelity}.
Photorealistic simulation data can be employed to train and validate visual perception algorithms, such as object detection, semantic segmentation, and SLAM.
Moreover, realistic models not only enhance visual fidelity but also improve interaction effects within simulations.
When robots perform actions such as grasping, collision, or force-based interactions, accurate geometry ensures stable and reliable feedback.
To achieve photorealistic appearance and minimize the sim-to-real visual gap, we addressed the following standards:

\textbf{Mesh.}
Manifold meshes form the core geometric foundation of each asset, defining the object's overall contour and spatial occupancy.
These meshes are critical for generating collision bodies that maintain accuracy in physical interactions.
\ours ensures that mesh details produce smooth surfaces and lifelike contours, avoiding jagged or blocky appearances.
Additionally, through normal vector optimization algorithms, redundant vertices are merged, reducing geometric data volume and thereby alleviating computational burdens in simulation.

\textbf{Texture.}
Textures are mapped onto mesh surfaces via UV coordinates to provide visual details.
\ours employs high-resolution textures to capture fine surface characteristics, such as the metallic sheen of a refrigerator or the subtle grain of wood on a chair.
Furthermore, textures are meticulously aligned with the UV map to prevent stretching, distortion, or visible seams.

\textbf{Material.}
A material is a collection of rendering parameters, including references to textures, that defines how an object's surface responds to light.
\ours leverages RTX Renderer~\citep{rtx} in Isaac Sim and adopts Physically Based Rendering (PBR)~\citep{pbr} to accurately simulate diffuse and specular reflections, enabling rendering effects such as roughness and emissive properties.
This approach allows for the realistic representation of diverse materials, achieving true-to-life visual fidelity.

\section{Physical Fidelity of Joints}
\label{subsec:Fidelity_of_Joints}
To achieve physical fidelity of dynamic joint and simulate variable joints motions in the real world, we enhance the joint drive equation originally provided by Isaac Sim:
\begin{equation}
\tau = K(q) \cdot \left(q - q_{\text{target}}(q)\right) + D \cdot \left(\dot{q} - \dot{q}_{\text{target}}(q)\right)
\end{equation}
where $\tau$ represents the force($F$) and torque($T$) applied to drive the joint, $q$ and $\dot{q}$ are the joint position and velocity, respectively,  $D$ donates damping, and $K$ donates stiffness.
While this equation can model basic joint motions, it fails to fully replicate complex dynamic joint motions in the real world.
For complex joints such as door closers and light switches, $\tau$ may vary with $q$ and $\dot{q}$.
To accommodate the above situations, we design functions of $q$ and $\dot{q}$.

\textbf{Impact from $\dot{q}$.}
Friction must be accounted for in simulation and cannot be modeled as a constant.
It imposes resistance to the force generated by the joint drive $\tau$, and we propose the following equation with three different conditions:
\begin{subequations}
  \begin{empheq}[left={F_{\text{friction}}(\dot{q}) = \empheqlbrace}]{align}
    -F_{\text{ext}} \label{eq:static} & \quad \dot{q} = 0 \text{ and } |F_{\text{ext}}| \leq \mu_s \cdot \left(|F| + |T|\right) \\
    -\mu_s \cdot \left(|F| + |T|\right) \cdot \operatorname{sign}(F_{\text{ext}}) \label{eq:breakaway} & \quad \dot{q} = 0 \text{ and } |F_{\text{ext}}| > \mu_s \cdot \left(|F| + |T|\right) \\
    -D \cdot \dot{q} \cdot \operatorname{sign}(\dot{q}) \label{eq:dynamic} & \quad \dot{q} \neq 0
  \end{empheq}
  \label{eq:friction}
\end{subequations}

We illustrate the friction from static friction, to maximum static friction, and finally to dynamic friction, corresponding to conditions from Eqn.~\eqref{eq:static} through Eqn.~\eqref{eq:dynamic}.
\( F_{\text{ext}} \) denotes the static friction.
The coefficient $u_s$ denotes the static friction coefficient, which can be configured in Isaac Sim via the \texttt{Joint Friction} parameter. The \textit{sign} function ensures that the frictional force is applied in the correct direction.

\textbf{Impact from $q$.} The latch release mechanism exemplifies the position-dependent joint drive, we analyze a button-actuated trash bin lid mechanism.
When the button is depressed, it triggers a linkage to retract the spring-loaded latch, enabling the lid to freely rotate under torsional spring torque to $q_{\text{upper\_bound}}$.
\begin{subequations}
  \begin{empheq}[left={q_{\text{target}}(q) = \empheqlbrace}]{align}
    q_{\text{upper\_bound}} & \quad \text{if } q > q_{\text{threshold}} \text{ and } S_{\text{open}} = 1 \label{eq:target_open} \\
    q_{\text{lower\_bound}} & \quad \text{if } q < q_{\text{threshold}} \text{ and } S_{\text{open}} = 0 \label{eq:target_close}
  \end{empheq}
  \label{eq:q_target}
\end{subequations}

We further investigate joint motion with abrupt stiffness variations, exemplified by refrigerator door closers and magnetic latching mechanisms.
To maintain static equilibrium in the stationary state, a high stiffness value \( k_{\text{high}} \) is employed.
When \( S_{\text{open}} = 1 \) (door opening phase), the stiffness progressively decreases with increasing \( q \).
Upon exceeding the critical position \( q_{\text{threshold}} \), the stiffness reaches its minimum \( k_{\text{low}} \), and the joint target position switches to \( q_{\text{upper\_bound}} \).
During door closure, as \( q \) approaches \( q_{\text{threshold}} \) from above, the target position abruptly transitions to \( q_{\text{lower\_bound}} \), accompanied by an exponential stiffness surge to rapidly complete closure, emulating commercial door closer dynamics. This behavior is formalized as:
\begin{subequations}
  \label{eq:stiffness_model}
  \begin{empheq}[left={K(q) = \empheqlbrace}]{align}
    k_{\text{high}}, 
      & \quad q = q_{\text{lower\_bound}} 
      \label{eq:stiffness_high} \\
    k_{\text{high}} - \alpha q, 
      & \quad \begin{aligned}
          &\text{if } q_{\text{lower\_bound}} < q \leq q_{\text{threshold}} \text{ and } S_{\text{open}}=1 
        \end{aligned} 
      \label{eq:stiffness_open} \\
    k_{\text{low}} + k_{\text{max}} e^{-\lambda q}, 
      & \quad \begin{aligned}
          &\text{if } q_{\text{lower\_bound}} < q \leq q_{\text{threshold}} \text{ and } S_{\text{open}}=0 
        \end{aligned} 
      \label{eq:stiffness_close} \\
    k_{\text{low}}, 
      & \quad q_{\text{threshold}} < q < q_{\text{upper\_bound}} 
      \label{eq:stiffness_low}
  \end{empheq}
\end{subequations}

\section{Chart Comparison with Existing Datasets}
\label{subsec:Chart_Comparison}
We present a detailed comparison of \ours with existing articulated-object datasets in Tab.~\ref{tab:chartcomparison}.

\begin{table}[htbp]
\centering
\caption{Detailed comparison with existing articulated-object datasets.}
\label{tab:chartcomparison}
\resizebox{\textwidth}{!}{%
\begin{tabular}{lcccccc}
\toprule
 & Articulated Assets & Prismatic Joints & Revolute Joints & Visual Realism & Physical Fidelity & Modular Interaction \\
\midrule
\ours & \totalNum & 2156 & 1809 & high & high & 394 \\
BEHAVIOR-1K & 545 & 318 & 819 & medium & low & None \\
PartNet-Mobility & 2347 & 7659 & 4312 & low & low & None \\
\bottomrule
\end{tabular}%
}
\end{table}

\section{Visual Realism Comparison}
\label{subsec:Visual_Comparison}
\begin{figure}
  \centering
  \includegraphics[width=1\linewidth]{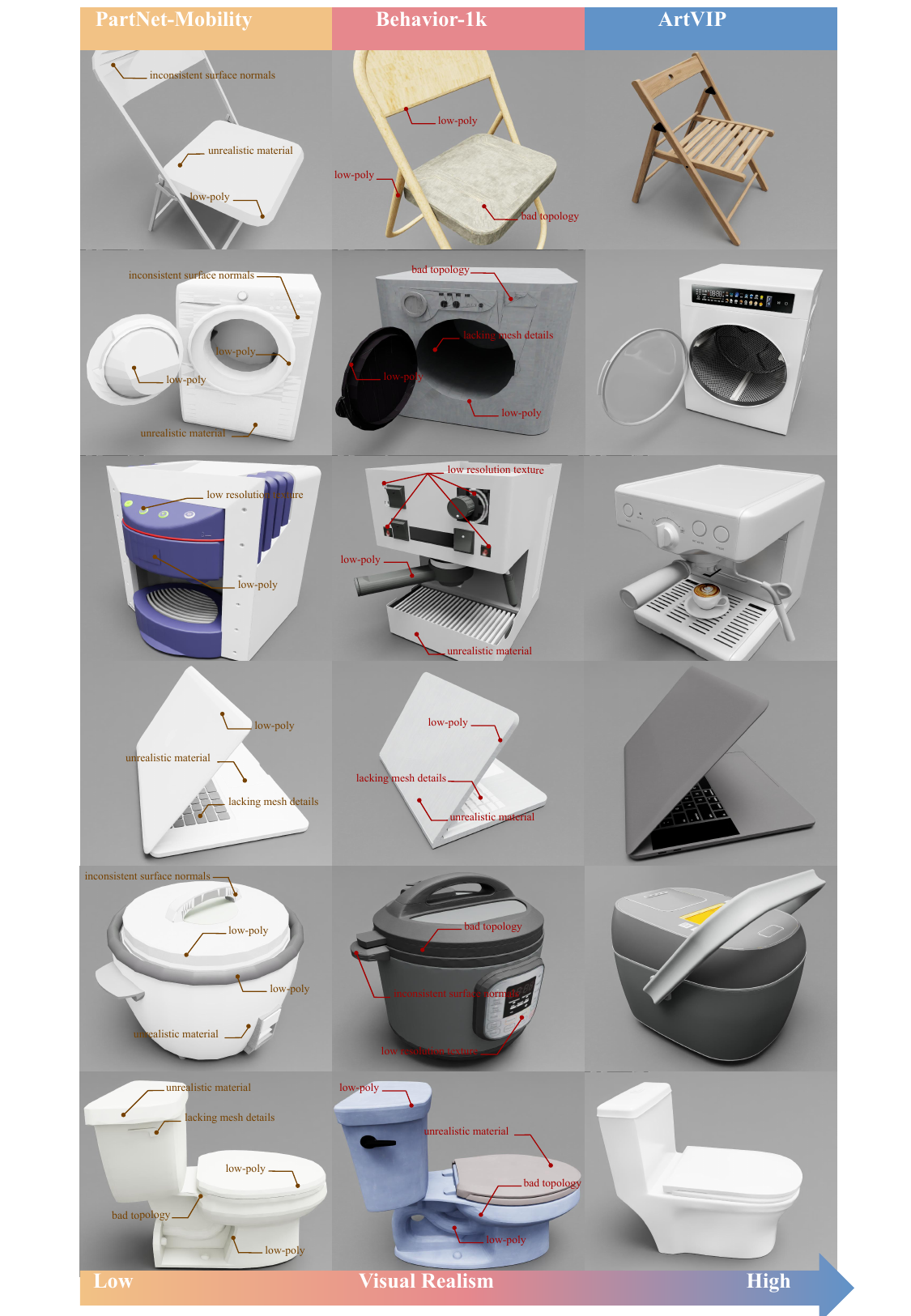}
  \caption{Comparisons of \ours, BEHAVIOR-1K, and PartNet-Mobility.}
  \label{fig:detail_compare}
\end{figure}

We present further comparative analysis in Fig.~\ref{fig:detail_compare}.
PartNet-Mobility employs the URDF format, with meshes stored in OBJ format and material information defined in MTL files.
Although the OBJ files are manually crafted, they frequently exhibit distorted meshes, significantly compromising visual quality.
The MTL material format inherently lacks the capability to model physically accurate light reflection, resulting in a lack of environmental realism across all PartNet-Mobility assets.
Our analysis reveals that many materials in PartNet-Mobility rely solely on base color for rendering, and the absence of textures substantially degrades the overall rendering quality.
Although BEHAVIOR-1K adopts the USD format, which supports physically based rendering (PBR), it still suffers from issues related to distorted meshes and poor texture quality.

To mitigate issues such as distorted meshes and angular surfaces, we employed a high number of triangular faces to ensure smooth surfaces and enhanced geometric detail.
For categories such as toilets and refrigerators, \ours significantly surpasses BEHAVIOR-1K and PartNet-Mobility in the number of triangular faces utilized. 
However, this approach entails a trade-off, as it reduces the simulation frame rate. To address this, we conducted profiling analysis to optimize the simulation frame rate for each object.
In our experiments, we selected the kitchen, which contains the highest number of articulated objects, and the living room, which features the most extensive texture rendering, as testing environments.
Each asset from \ours was individually placed within these scenes, ensuring that the overall rendering frame rate consistently exceeds 60~Hz (i7-13700, Nvidia 4090, 64~GB).

To study the effect of triangle count, we report comprehensive statistics in Tab.~\ref{tab:category_averages} for ArtVIP, PartNet-Mobility, and BEHAVIOR-1K: the average triangle count, the average number of active joints, the average FPS with a single asset, and the average FPS in the kitchen scene. The kitchen scene is the most complex environment, containing 65 actuated joints.
ArtVIP and PartNet-Mobility are evaluated in Isaac Sim 5.1. BEHAVIOR-1K assets are encrypted and accessible only through OmniGibson (Isaac Sim 4.5). We attribute the large FPS fluctuations observed for BEHAVIOR-1K to overhead introduced by the derivative framework.
Based on the FPS results for ArtVIP and PartNet-Mobility, we conclude:
1) For a single object, under Isaac Sim's iterative optimizations, triangle counts up to approximately 100k and up to 20 active joints have negligible impact on FPS.
2) In complex scenes, both triangle count and the number of active joints reduce FPS.

\begin{table}[t]
\centering
\small
\caption{Category-wise averages for triangle count, active joints, and FPS across datasets.}
\label{tab:category_averages}
\resizebox{\textwidth}{!}{
\begin{tabular}{lrrrrrrrrrrrr}
\toprule
Item Category & \multicolumn{3}{c}{Avg.\ Triangle Count} & \multicolumn{3}{c}{Avg.\ Active Joints} & \multicolumn{3}{c}{Avg.\ FPS (Single Item)} & \multicolumn{3}{c}{Avg.\ FPS (In Kitchen)} \\
\cmidrule(lr){2-4} \cmidrule(lr){5-7} \cmidrule(lr){8-10} \cmidrule(lr){11-13}
 & ArtVIP & PartNet-Mobility & BEHAVIOR-1K & ArtVIP & PartNet-Mobility & BEHAVIOR-1K & ArtVIP & PartNet-Mobility & BEHAVIOR-1K & ArtVIP & PartNet-Mobility & BEHAVIOR-1K \\
\midrule
Coffee Machine & 80484.8 & 27104.7 & 42256 & 2.2 & 5.759 & 5.5 & 91.97 & 91.95 & 114.77 & 72.01 & 73.83 & 48.33 \\
Microwave & 34494.6 & 8620.5 & 5521 & 4 & 4.313 & 1.857 & 91.93 & 91.88 & 87.51 & 69.38 & 72.95 & 50.04 \\
Oven & 99048 & 41206.2 & 25638 & 4.5 & 6.133 & 1 & 91.98 & 91.95 & 109.97 & 72.66 & 74.64 & 49.4 \\
Dishwasher & 54427.1 & 8932.6 & 25162.2 & 1.429 & 1.333 & 2.5 & 91.94 & 91.95 & 115.2 & 65.75 & 69.95 & 49.73 \\
Rice Cooker & 101573.3 & 26068.7 & 40245.3 & 3.333 & 1.12 & 1 & 91.96 & 91.96 & 115.58 & 70.97 & 74.91 & 48.29 \\
Laptop & 46053.6 & 37378.7 & 18546.3 & 1 & 1 & 1 & 91.97 & 91.93 & 112 & 74.59 & 74.89 & 46.99 \\
Washing Machine & 151705.4 & 26269.8 & 27380.8 & 2.57 & 7.471 & 1.538 & 91.95 & 91.94 & 107.17 & 70.97 & 70.74 & 48.27 \\
Toilet & 164271.6 & 22276.49 & 15011.11 & 3.6 & 2.319 & 2.611 & 91.95 & 91.95 & 120.58 & 74.94 & 74.93 & 47.18 \\
Refrigerator & 100903.8 & 6517 & 24273.4 & 6.25 & 1.682 & 1.538 & 91.94 & 91.96 & 99.55 & 60.78 & 73.89 & 49.82 \\
Table & 20184.7 & 22607.6 & 14210.6 & 5.28 & 3.158 & 2.633 & 91.96 & 91.96 & 116.37 & 68.13 & 71.68 & 47.92 \\
Folding Chair & 21567.5 & 6519.3 & 7064.6 & 2 & 1.231 & 2 & 91.95 & 91.97 & 125.93 & 74.5 & 74.92 & 51.41 \\
Scissors & 43953 & 14601 & 4972 & 2 & 1.963 & 1 & 91.96 & 91.96 & 129.71 & 72.5 & 74.92 & 52.45 \\
Trash Can & 30139.6 & 6468.33 & 8370.17 & 1.77 & 1.971 & 1 & 91.94 & 91.93 & 121.28 & 71.66 & 74.93 & 52.24 \\
\bottomrule
\end{tabular}
}
\end{table}

\section{Physical Fidelity and Interaction Evaluations}
\label{subsec:Physical_Evaluations}
\begin{figure}
  \centering
  \includegraphics[width=1\linewidth]{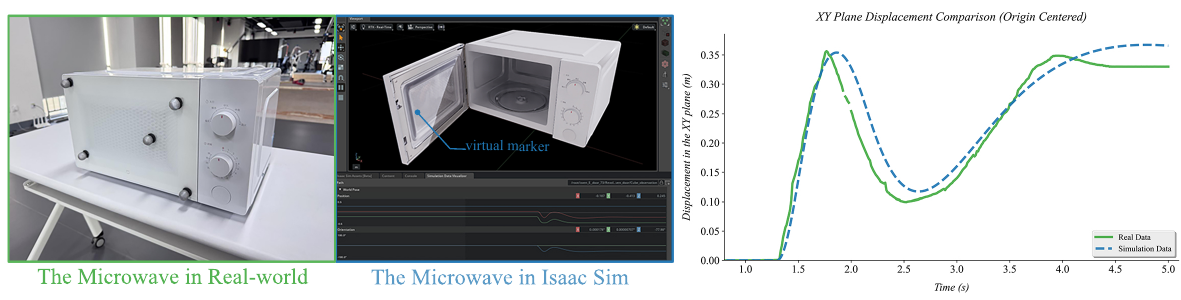}
  \caption{
    \textbf{Left and Middle:} Digital-twin asset examples in real-world and simulation.
    \textbf{Right:} Analysis of the Microwave's displacement.
  }
  \label{fig:mocap_micro}
\end{figure}

\textbf{Motion Triggered by Latch Release.}
To validate the modular interaction within assets, we compared the triggered joint in both real-world and virtual microwave.
We conducted button-press experiments in each environment to initiate the door-opening action and recorded the resulting door motion trajectories.
In the real-world tests we tracked a marker on the door using the optical tracking system to capture its spatial motion after the button pressed.
In the simulation we set a virtual marker at the same position as the real-world marker on the door, and we triggered the door opening via pressing the button as well (for which the activation configured in modular interaction) and logged the virtual marker's trajectories.
We performed ten trials in each environment and computed the average spatial trajectory as Fig.~\ref{fig:mocap_micro} shown.

\begin{figure}
  \centering
  \includegraphics[width=1\linewidth]{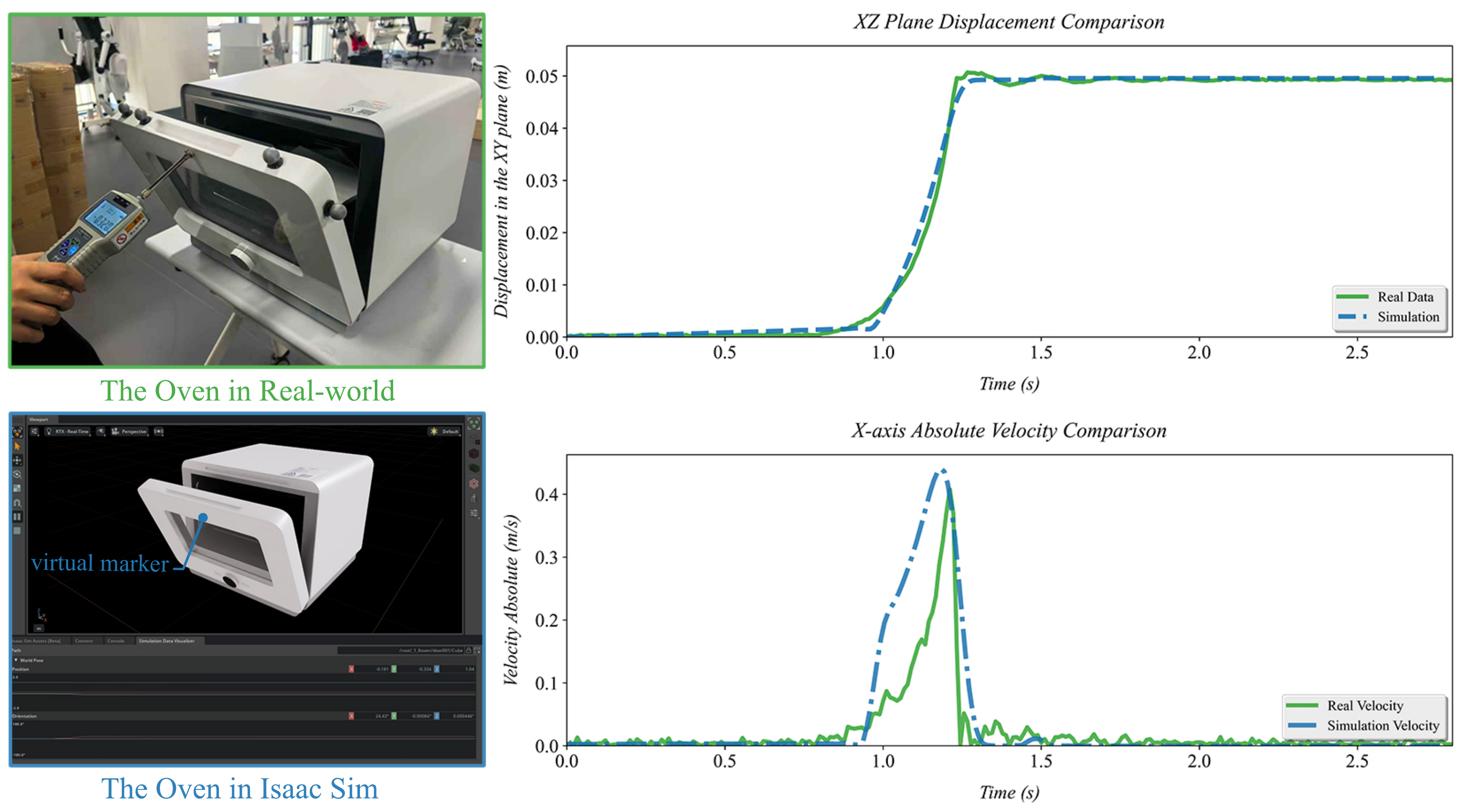}
  \caption{
    \textbf{Left:} Digital-twin asset examples in real-world and simulation.
    \textbf{Right:} Analysis of the oven's displacement.
  }
  \label{fig:mocap_oven}
\end{figure}

\textbf{Motion Triggered by Joint Position Threshold.}
Appliances equipped with door closers typically exhibit a dynamic change in motion once the door reaches a certain angle during closing.
After arriving at a certain angle, the door closer causes the door to accelerate and snap shut against the appliance body.
To evaluate how well the simulation captures this physical transition, we focus on analyzing the door's linear and angular velocities during the transition from the threshold state to full closure.
In both the simulation and real-world experiments, a force of no more than $1.0$ N is applied when the door is within the threshold range to trigger the door closer mechanism.
We then record the kinematic behavior following the activation of the door closer.
In the real-world setup, the optical motion capture system is used to track the spatial displacement of markers on the door.
Both the simulation and real-world experiments are repeated ten times, and we compute the average spatial trajectories and changes in velocity along the X-axis for quantitative comparison (Fig.~\ref{fig:mocap_oven}).

\section{Imitation Learning Application}
\label{subsec:IL_APP}
\begin{figure}[tp]
  \centerline{\includegraphics[width=0.99\linewidth]{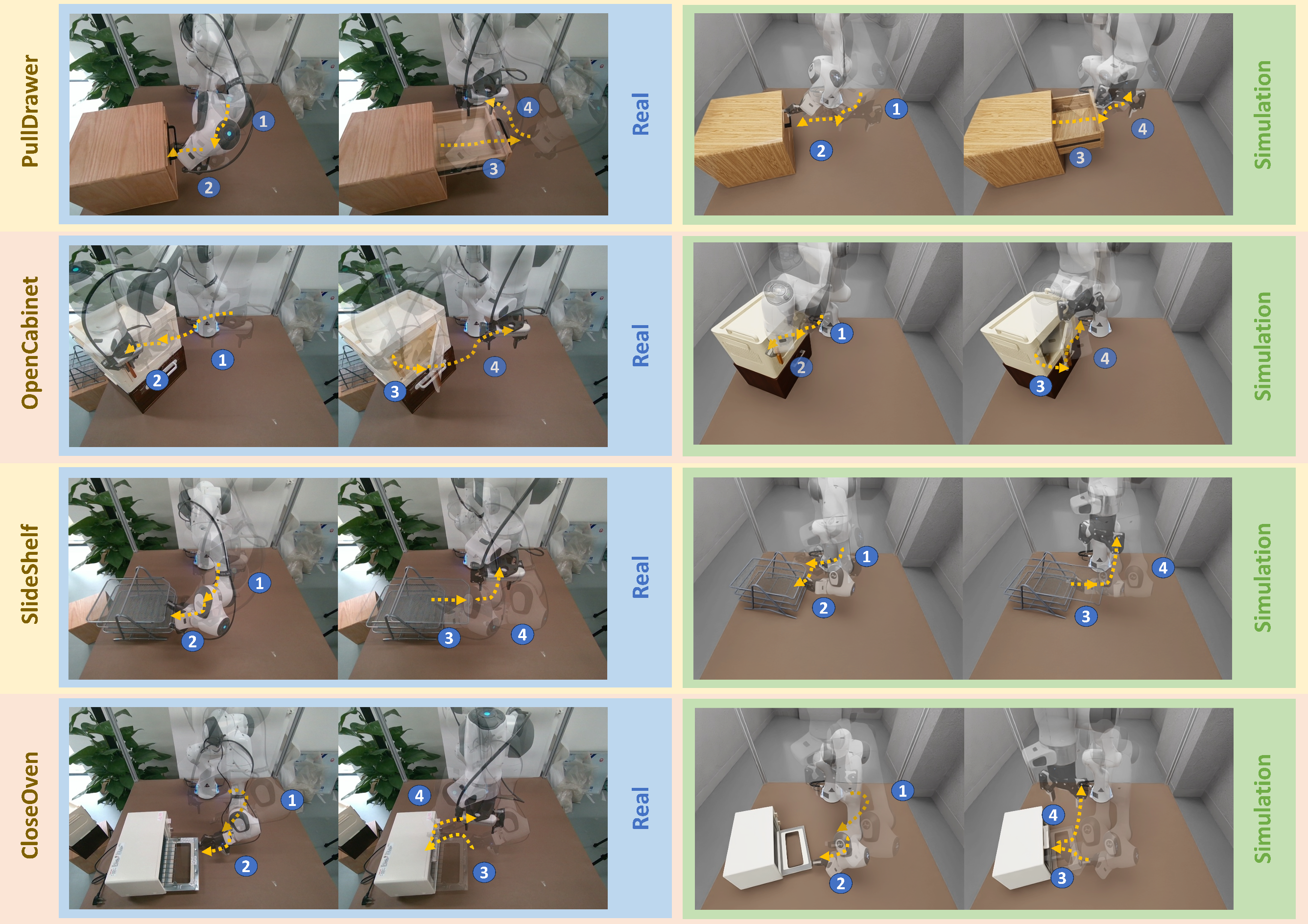}}
  \caption{The four articulated-object manipulation tasks conducted for imitation learning.}
  \label{fig:il_task_summary}
\end{figure}

\textbf{Task Summary.}
As shown in Fig.~\ref{fig:il_task_summary}, we design four challenging articulated-object manipulation tasks: 
(1) \textbf{PullDrawer}, 
(2) \textbf{OpenCabinet}, 
(3) \textbf{SlideShelf}, and 
(4) \textbf{CloseOven}.
These tasks demand precise and flexible motions, including rotation, angled pushing, and horizontal translation.
We define these tasks as follows: 
\begin{itemize}
  \item 
    \textbf{PullDrawer}. 
    his task requires the robot to insert the gripper into the handle of the drawer, securely press the handle, and gradually pull the drawer out along a linear trajectory using a smooth and consistent motion. 
  \item 
    \textbf{OpenCabinet}. 
    For this task, the robotic arm needs to precisely locate the thin vertical handle of the cabinet door. 
    The gripper has to align vertically, firmly grip the handle, and pull the door outward along a curved path while maintaining a stable trajectory.  
  \item 
    \textbf{SlideShelf}.
    This task involves horizontal manipulation of the shelf.
    First, the gripper needs to rotate around $90$ degrees to align parallel to the shelf's direction.
    It then grips the base of the shelf and moves horizontally, pulling the shelf out along its guide rails in a stable and controlled manner.  
  \item 
    \textbf{CloseOven}.
    To complete this task, the robotic arm needs to close its gripper to push against the bottom edge of the oven door.
    The arm then rotates and lifts under the door, applying a curved upward force to close the door. 
\end{itemize}

\textbf{Imitation Learning Algorithm.}
The input to the imitation learning models consists of RGB image data from multiple camera views and the robot's proprioceptive states.
The output is the robot control signals, such as joint positions, enabling end-to-end task execution. 
We used two state-of-the-art imitation learning methods, Action Chunking Transformer (ACT)~\citep{zhao2023learningfinegrainedbimanualmanipulation} and Diffusion Policy (DP)~\citep{chi2023diffusion}, to train the robotic policies for the articulated object manipulation task. Hyperparameters of both methods are demonstrated in Tab.~\ref{tab:para_act} and Tab.~\ref{tab:para_dp}.
\begin{itemize}
    \item \textbf{Action Chunking Transformer (ACT)}~\citep{zhao2023learningfinegrainedbimanualmanipulation}: ACT is built on the transformer network architecture and leverages temporal ensemble techniques to produce fluid and precise action sequences.
    \item \textbf{Diffusion Policy (DP)}~\citep{chi2023diffusion}: DP employs a diffusion-based generative model that captures multi-modal action distributions, offering robustness and high success rates for complex robotic tasks.
\end{itemize}

\textbf{Detailed Experiment Results.} The Full experiment results are presented in Tab.~\ref{tab:significance_test_a}. 

\begin{table}[t]
  \centering
  \resizebox{\columnwidth}{!}{
      \begin{tabular}{cllcll}
      \toprule
      \multirow{7}{*}{Training} & Hyperparameter & Value &  \multirow{7}{*}{\shortstack{Network\\Architectures}} & Hyperparameter & Value \\
      \cmidrule(lr){2-3} \cmidrule(lr){5-6} 
      & Batch size & 48 & & Encoder layer & 4 \\
      & Learning rate & 1e-4 & & Decoder layer & 7 \\ 
      & Optimizer & AdamW & & Forward dim & 3200 \\ 
      & KL weight & 10 & & Heads num & 8 \\ 
      & Action sequence & 50 & & Transformer hidden dim & 512 \\ 
      & Training step & 50k & & Backbone & ResNet50 \\ 
      \bottomrule
      \end{tabular}
  }
  \caption{Implementation details of Action Chunking Transformer (ACT).}
  \label{tab:para_act}
\end{table}

\begin{table}[t]
\centering
\resizebox{\columnwidth}{!}{
    \begin{tabular}{cllcll}
    \toprule
    \multirow{7}{*}{Training} & Hyperparameter & Value & \multirow{7}{*}{\shortstack{Network\\Architectures}} & Hyperparameter & Value \\
    \midrule
    & Batch size & 48 &  & Diffuion Network & Unet1D \\
    & Learning rate & 1e-4 & & Pooling & SpatialSoftmax \\ 
    & Optimizer & AdamW & & Noise scheduler & DDIM \\ 
    & EMA power & 0.75 & & EMA model & True \\ 
    & Action sequence & 16 & & Noise schedule & SquaredcosCap \\ 
    & Training step & 50k & & Backbone & ResNet50 \\ 
    \bottomrule
    \end{tabular}
}
\caption{Implementation details of Diffusion Policy (DP).}
\label{tab:para_dp}
\end{table}

\begin{table}[t]
\centering
\caption{Performance results (scheme A): per-seed scores and mean $\pm$ 90\% CI}
\label{tab:significance_test_a}
\footnotesize
\begin{center}
\begin{tabular}{lllcccc}
\multicolumn{1}{c}{\textbf{Task}} & \multicolumn{1}{c}{\textbf{Method}} & \multicolumn{1}{c}{\textbf{Strategy}} & \multicolumn{1}{c}{\textbf{Seed 1}} & \multicolumn{1}{c}{\textbf{Seed 2}} & \multicolumn{1}{c}{\textbf{Seed 3}} & \multicolumn{1}{c}{\textbf{Mean $\pm$ CI$_{90}$}} \\
\hline 
\multirow{12}{*}{PullDrawer} & \multirow{6}{*}{ACT} & RO & 0.567 & 0.767 & 0.600 & 0.644 $\pm$ 0.059 \\
& & SO & 0.433 & 0.433 & 0.300 & 0.389 $\pm$ 0.060 \\
& & RSM100+10 & 0.500 & 0.667 & 0.767 & 0.640 $\pm$ 0.059 \\
& & RSM100+20 & 0.667 & 0.600 & 0.767 & 0.678 $\pm$ 0.057 \\
& & RSM100+50 & 0.833 & 0.767 & 0.733 & 0.778 $\pm$ 0.051 \\
& & RSM100+100 & 0.767 & 0.867 & 0.800 & 0.811 $\pm$ 0.048 \\
\cline{2-7}
& \multirow{6}{*}{DP} & RO & 0.600 & 0.733 & 0.633 & 0.656 $\pm$ 0.058 \\
& & SO & 0.133 & 0.233 & 0.233 & 0.200 $\pm$ 0.049 \\
& & RSM100+10 & 0.600 & 0.650 & 0.700 & 0.650 $\pm$ 0.057 \\
& & RSM100+20 & 0.650 & 0.700 & 0.733 & 0.694 $\pm$ 0.056 \\
& & RSM100+50 & 0.700 & 0.733 & 0.750 & 0.728 $\pm$ 0.055 \\
& & RSM100+100 & 0.733 & 0.767 & 0.867 & 0.789 $\pm$ 0.050 \\
\hline
\multirow{12}{*}{OpenCabinet} & \multirow{6}{*}{ACT} & RO & 0.300 & 0.400 & 0.333 & 0.344 $\pm$ 0.058 \\
& & SO & 0.167 & 0.100 & 0.100 & 0.122 $\pm$ 0.040 \\
& & RSM100+10 & 0.333 & 0.367 & 0.367 & 0.356 $\pm$ 0.059 \\
& & RSM100+20 & 0.367 & 0.400 & 0.367 & 0.378 $\pm$ 0.059 \\
& & RSM100+50 & 0.433 & 0.500 & 0.400 & 0.444 $\pm$ 0.061 \\
& & RSM100+100 & 0.567 & 0.367 & 0.433 & 0.456 $\pm$ 0.061 \\
\cline{2-7}
& \multirow{6}{*}{DP} & RO & 0.467 & 0.500 & 0.500 & 0.489 $\pm$ 0.061 \\
& & SO & 0.133 & 0.033 & 0.133 & 0.100 $\pm$ 0.037 \\
& & RSM100+10 & 0.500 & 0.533 & 0.567 & 0.533 $\pm$ 0.058 \\
& & RSM100+20 & 0.550 & 0.583 & 0.600 & 0.578 $\pm$ 0.057 \\
& & RSM100+50 & 0.600 & 0.617 & 0.633 & 0.617 $\pm$ 0.057 \\
& & RSM100+100 & 0.667 & 0.700 & 0.600 & 0.656 $\pm$ 0.058 \\
\hline
\multirow{12}{*}{SlideShelf} & \multirow{6}{*}{ACT} & RO & 0.233 & 0.233 & 0.333 & 0.267 $\pm$ 0.054 \\
& & SO & 0.100 & 0.167 & 0.133 & 0.133 $\pm$ 0.042 \\
& & RSM100+10 & 0.200 & 0.267 & 0.300 & 0.256 $\pm$ 0.053 \\
& & RSM100+20 & 0.233 & 0.300 & 0.267 & 0.267 $\pm$ 0.054 \\
& & RSM100+50 & 0.300 & 0.367 & 0.300 & 0.322 $\pm$ 0.057 \\
& & RSM100+100 & 0.333 & 0.333 & 0.400 & 0.356 $\pm$ 0.059 \\
\cline{2-7}
& \multirow{6}{*}{DP} & RO & 0.467 & 0.433 & 0.433 & 0.444 $\pm$ 0.061 \\
& & SO & 0.167 & 0.167 & 0.200 & 0.178 $\pm$ 0.047 \\
& & RSM100+10 & 0.433 & 0.467 & 0.500 & 0.467 $\pm$ 0.058 \\
& & RSM100+20 & 0.500 & 0.533 & 0.550 & 0.528 $\pm$ 0.057 \\
& & RSM100+50 & 0.533 & 0.567 & 0.583 & 0.561 $\pm$ 0.056 \\
& & RSM100+100 & 0.567 & 0.600 & 0.600 & 0.589 $\pm$ 0.060 \\
\hline
\multirow{12}{*}{CloseOven} & \multirow{6}{*}{ACT} & RO & 0.500 & 0.633 & 0.600 & 0.578 $\pm$ 0.061 \\
& & SO & 0.267 & 0.267 & 0.167 & 0.233 $\pm$ 0.052 \\
& & RSM100+10 & 0.500 & 0.600 & 0.667 & 0.589 $\pm$ 0.060 \\
& & RSM100+20 & 0.533 & 0.633 & 0.633 & 0.600 $\pm$ 0.060 \\
& & RSM100+50 & 0.733 & 0.533 & 0.700 & 0.656 $\pm$ 0.058 \\
& & RSM100+100 & 0.667 & 0.800 & 0.567 & 0.678 $\pm$ 0.057 \\
\cline{2-7}
& \multirow{6}{*}{DP} & RO & 0.600 & 0.700 & 0.667 & 0.656 $\pm$ 0.058 \\
& & SO & 0.267 & 0.233 & 0.333 & 0.278 $\pm$ 0.055 \\
& & RSM100+10 & 0.633 & 0.667 & 0.700 & 0.667 $\pm$ 0.057 \\
& & RSM100+20 & 0.667 & 0.700 & 0.733 & 0.700 $\pm$ 0.056 \\
& & RSM100+50 & 0.700 & 0.733 & 0.750 & 0.728 $\pm$ 0.055 \\
& & RSM100+100 & 0.767 & 0.733 & 0.833 & 0.778 $\pm$ 0.051 \\
\hline
\end{tabular}
\end{center}
\end{table}

\section{Reinforcement Learning Application}
\label{subsec:RL_APP}
\begin{figure*}[t]
  \centering
  \begin{subfigure}[t]{0.48\textwidth}
    \centering
    \includegraphics[width=\linewidth]{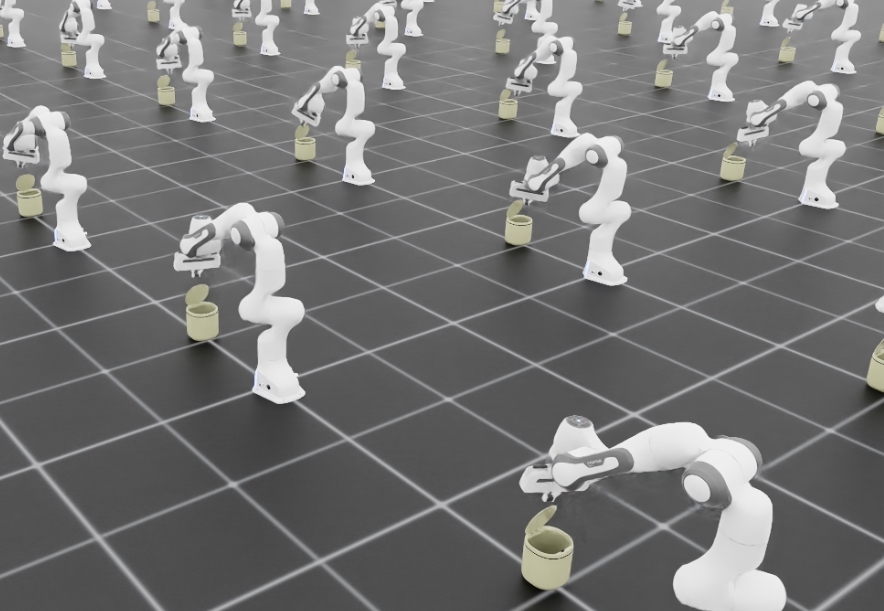}
    \caption{Training task.}
    \label{fig:drl-env}
  \end{subfigure}
  \hspace{10pt}
  \begin{subfigure}[t]{0.42\textwidth}
    \centering
    \includegraphics[width=\linewidth]{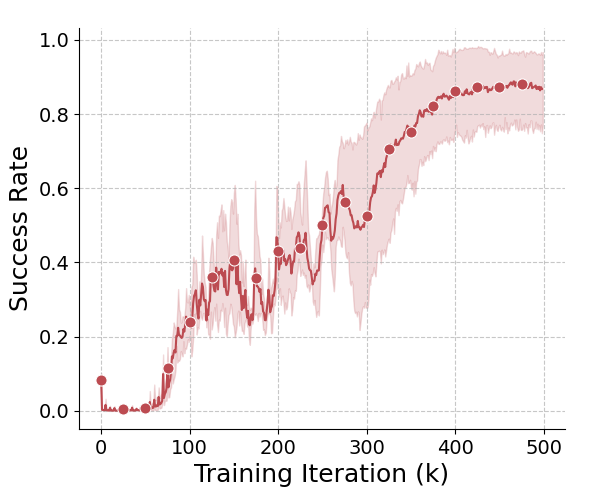}
    \caption{Training curve over five random seeds.}
    \label{fig:drl-training}
  \end{subfigure}
  \caption{RL-based training of visuomotor policy with \ours.}
  \label{fig:training}
\end{figure*}

\textbf{Training Details.}
We extend the visual RL framework EAGLE~\citep{zhao2025efficient} to articulated-object tasks in \ours. Fig.~\ref{fig:drl-env} shows the CloseTrashcan task, where the robot arm is required to close the trashcan within a given time limit.
EAGLE is a two-stage visual RL framework designed for efficiency and generalization.
In Stage 1, the teacher policy receives low-level states, including the robot arm's proprioceptive input, the lid's joint value, and the 3D relative position between the trashbin and the gripper.
In Stage 2, the student policy is provided only with the wrist camera image and the robot's proprioceptive state—no object-related states are available. Fig.~\ref{fig:drl-training} presents the training curves in Stage 2.

For implementation details, in Stage 1, we replace EAGLE's original RL agent with PPO; In Stage 2, a privileged-state teacher is distilled into a visuomotor student while a self-supervised attention mask learned as follows:
\begin{equation}
  \label{eqn:att_loss}
  \mathcal{L}_{att}=\mathcal{L}_{rec}+\mathcal{L}_{ae}+\beta\mathcal{L}_{ctl}+\lambda\mathcal{L}_{sps},
\end{equation}
where $\mathcal{L}_{rec}$ and $\mathcal{L}_{ae}$ are reconstruction losses, $\mathcal{L}_{ctl}$ predicts dynamics, and $\mathcal{L}_{sps}$ enforces mask sparsity.
Hyper-parameters $\beta$ and $\lambda$ weight auxiliary losses.

The student policy is trained with the distillation loss:
\begin{equation}
  \label{eqn:cloning_loss}
  \hat{\mathcal{L}}(\pi_{\theta})=\mathbb{E}_{(\mathbf{o},\mathbf{s})\sim\mathcal{D}}
  \bigl[\lVert\pi_{\theta}(\mathbf{o}_{\text{aug}})-\pi_{e}(\mathbf{s})\rVert_2^{2}\bigr],
\end{equation}
where $\mathbf{s}$ contains privileged states and $\mathbf{o}_{\text{aug}}$ are images augmented by the learned mask with Eqn.~\eqref{eqn:att_loss}.
Hyper-parameters used in EAGLE are listed in Tab.~\ref{tab:para_eagle}.

\begin{table}[h]
\centering
\resizebox{0.8\columnwidth}{!}{
    \begin{tabular}{cll}
    \toprule
     & Hyperparameter & Value \\ 
    \midrule
    \multirow{6}{*}{\makecell{Teacher \\ (Stage 1)}} 
    & Learning rate for all net & 5e-4 \\
    & Optimizer & Adam \\ 
    & Batch size & 12 $\times$ 4096 \\ 
    & Discount factor & 0.99 \\ 
    & Clip ratio & 0.2 \\ 
    & Rollout size & 96 $\times$ 4096 \\
    \cmidrule(lr){1-3}
    \multirow{9}{*}{\makecell{Student \\ (Stage 2)}}
    & Observation & 128 $\times$ 128 \\ 
    & Learning rate for all net & 1e-4 \\ 
    & Optimizer & Adam \\
    & Batch size & 256 \\ 
    & Frame stack & 1 \\ 
    & Replay buffer size & 100k \\
    & $\lambda$ & 0.01 \\
    & $\beta$ & 0.5 \\
    & $\alpha$ in $\textit{random overlay}$ & linear schedule from 0.4 to 0.9 \\  
    \bottomrule
    \end{tabular}
}
\caption{Hyperparamters for EAGLE.}
\label{tab:para_eagle}
\end{table}

\textbf{Reward Functions.}
The \textbf{CloseTrashcan} task is a long-horizon challenge requiring the robot to first approach the trashcan lid and then close it smoothly. To facilitate efficient RL training, we design a multi-objective reward function as follows:
\begin{equation}
  r_t(\bm{s}_t, \bm{a}_t) = \lambda_1 r_{dst}(\bm{s}_t) + \lambda_2 r_{dir}(\bm{s}_t) + \lambda_3 r_{cls}(\bm{s}_t) + \lambda_4 r_{smth} (\bm{a}_t),
\end{equation}
where $r_{dst}$ rewards proximity between the gripper and the lid, $r_{dir}$ encourages alignment toward the lid, $r_{cls}$ measures lid closure progress, and $r_{smth}$ promotes smooth actions.
The reward weights are set as: $\lambda_1 = 0.5$, $\lambda_2 = 0.125$, $\lambda_3 = 10$, $\lambda_4 = -0.01$.

\textbf{Baseline Comparison.}
To put EAGLE’s performance in context, we compare it with a vision-based PPO method. As shown in Tab.~\ref{tab:rl_eagle_vs_ppo}, due to the high computation complexity and low data diversity, the baseline performs poorly on the CloseTrashcan task, while EAGLE achieves a 98\% success rate after 500k training iterations.

\begin{table}[h]
\centering
\small
\caption{EAGLE vs. vision-based PPO: success rate across training checkpoints (k).}
\label{tab:rl_eagle_vs_ppo}
\begin{tabular}{lccccc}
  \toprule
  \multirow{2}{*}{\textbf{Method}} & \multicolumn{5}{c}{\textbf{Training Iterations (k)}} \\
  \cmidrule(lr){2-6}
   & 100 & 200 & 300 & 400 & 500 \\
  \midrule
  EAGLE & 0.23 & 0.28 & 0.73 & 0.85 & 0.98 \\
  Vision-based PPO & 0.16 & 0.19 & 0.21 & 0.22 & 0.24 \\
  \bottomrule
\end{tabular}
\end{table}

\section{Pearson correlation coefficient details}
\label{subsec:Pearson}
Following~\citep{li2024evaluatingrealworldrobotmanipulation}, we compute the Pearson correlation coefficient from the success rates in Tab.~\ref{tab:rl_sim_vs_real} as
$$r = \frac{\sum_{i=1}^n (x_i - \bar{x})(y_i - \bar{y})}{\sqrt{\sum_{i=1}^n (x_i - \bar{x})^2}\sqrt{\sum_{i=1}^n (y_i - \bar{y})^2}},$$
where $x_i$ and $y_i$ denote the corresponding success rates in simulation and the real world at the $i$-th checkpoint. A high Pearson correlation indicates a strong linear relationship between simulated and real-world performance. The value $r = 0.9886$ using data from Tab.~\ref{tab:rl_sim_vs_real} shows that ArtVIP provides a reliable simulated training and evaluation pipeline for RL.

\section{Comparison to Generative Pipelines}
\label{subsec:generative_pipeline}
\begin{figure}[!b]
  \centering
  \includegraphics[width=0.95\linewidth]{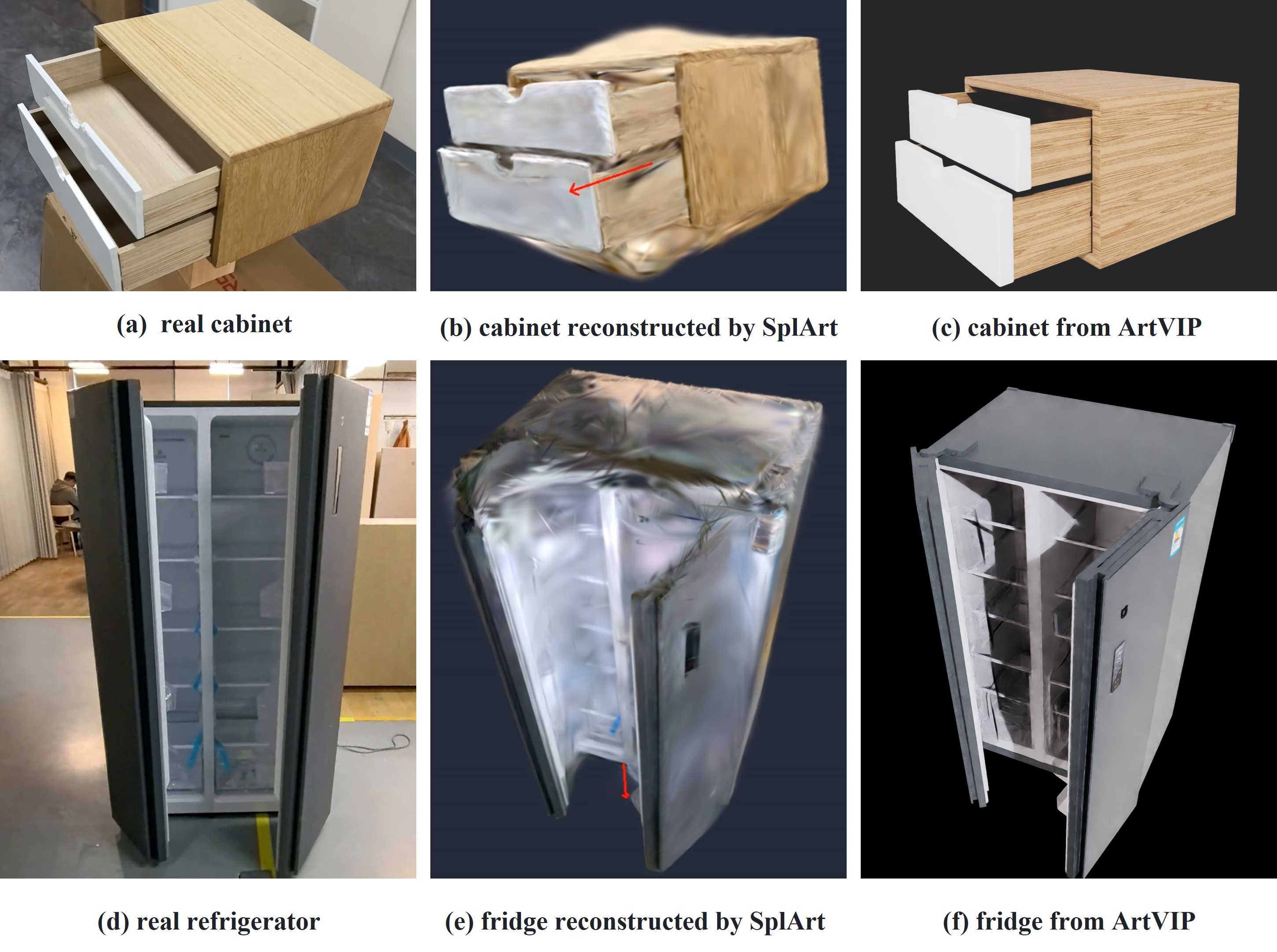}
  \caption{Comparison of real-world objects, generated outputs, and digital-twin assets.}
  \label{fig:generative_pic}
\end{figure}

To evaluate the quality of generated assets, we reproduced SplArt~\citep{lin2025splart} and generated a two\textendash drawer cabinet and a side\textendash by\textendash side fridge. As shown in Fig.~\ref{fig:generative_pic}, the outputs are of lower quality than our digital\textendash twin assets. Representative failures include 1) self‑collisions, 2) severe mesh distortions and breakage, 3) incorrect joint limits, positions, and axes, 4) materials and colors that deviate markedly from reality, and 5) severe lack of interior details. Consequently, current generative baselines fail to produce simulation\textendash ready articulated assets. 

To quantify the domain gap, we evaluate reconstruction metrics on two real\textendash world objects (cabinet and fridge) and compare them with the metrics reported by SplArt~\citep{lin2025splart} for the same categories on synthetic data (Tab.~\ref{tab:artgs_metrics}). While SplArt achieves low errors on synthetic inputs, performance degrades markedly on real\textendash world data, highlighting a substantial sim\textendash to\textendash real gap in articulated reconstruction. Additionally, the generated cabinet and fridge contain $\sim$100$\times$ more triangles than their \ours counterparts, reducing the simulation frame rate from $\sim$90 fps to $\sim$70 fps.

Limitations of generative pipelines:
Many baselines~\citep{lin2025splart,mandi2024real2codereconstructarticulatedobjects,liu2025artgsbuildinginteractablereplicas,guo2025articulatedgs,liu2024singapo,su2024artformer} are primarily trained and evaluated on synthetic data, where calibrated camera poses and controllable materials reduce domain gap. When performing inference on real images, they face:
\begin{enumerate}
  \item Reconstruction accuracy degrades when camera extrinsics are estimated rather than known.
  \item Since these methods are trained on PartNet\textendash Mobility, generalization is constrained by that dataset’s category coverage and articulation priors.
  \item Flat surfaces and right\textendash angle structures often appear wavy or warped under monocular/low\textendash texture settings (e.g., Real2Code~\citep{mandi2024real2codereconstructarticulatedobjects}, Fig.~5).
  \item Reconstruction outputs frequently lack physically consistent materials and textures. 3DGS methods encode color via spherical harmonics (f\_dc/f\_rest) requiring custom shaders, so rendering\textendash pipeline simulators (Isaac Sim, Sapien, PyBullet, MuJoCo, etc.) cannot render colors correctly without conversion/baking.
  \item Current metrics do not capture mesh/triangle efficiency or collision performance relevant to simulation FPS.
\end{enumerate}

\begin{table}[t]
  \centering
  \caption{Real-world vs.\ SplArt (synthetic) reconstruction metrics for cabinet and fridge.}
  \label{tab:artgs_metrics}
  \begin{tabular}{lcccc}
    \toprule
    Metric & \multicolumn{2}{c}{Real-World} & \multicolumn{2}{c}{SplArt (Synthetic)} \\
    \cmidrule(lr){2-3} \cmidrule(lr){4-5}
     & Cabinet & Fridge & Cabinet & Fridge \\
    \midrule
    CD-Static & 13.71 & 11.04 & 7.31 & 0.52 \\
    CD-Mobile & 11.50 & 13.45 & 1.02 & 0.27 \\
    CD-Whole & 12.35 & 12.78 & 5.21 & 0.70 \\
    Axis Ang. & 20.90 & N/A & 0.01 & 0.03 \\
    Axis Pos. & 0.094 & N/A & -- & 0.00 \\
    \bottomrule
  \end{tabular}
  \caption*{Notes: CD denotes Chamfer Distance ($\downarrow$ better). Axis Angular Error and Axis Positional Error are of the joint axis ($\downarrow$ better). ``N/A'' indicates the axis estimate cannot be produced (e.g., joint not detected).}
\end{table}

\newpage
\section{The Use of Large Language Models.}
A large language model (LLM) was used strictly as a writing aid for language polishing (grammar, clarity, and style). All ideas, methodological designs, datasets, code, analyses, and results are original and solely produced by the authors.

\end{document}